\documentclass[journal]{IEEEtran}

\ifCLASSOPTIONcompsoc
\usepackage[nocompress]{cite}
\else
\usepackage{cite}
\fi
\ifCLASSINFOpdf
\else
\fi
\usepackage{times}
\usepackage{epsfig}
\usepackage{graphicx}
\usepackage{amsmath}
\usepackage{amssymb}
\usepackage{bm}

\usepackage{mathtools}
\usepackage{tikz}
\usetikzlibrary{trees}

\usepackage{pifont}
\usepackage{multirow, makecell}

\usepackage{comment}

\usepackage{dblfloatfix}

\usetikzlibrary{arrows.meta, shapes.geometric, calc, shadows}
\colorlet{linecol}{black!75}

\usepackage{forest}
\usepackage{newfloat}
\DeclareFloatingEnvironment[fileext=lod]{Figure} %for adding captions

\forestset{/.style=
{
	{for tree=
		{parent anchor=south, 
			child anchor=north,
			align=center
			l+=1.5cm}
	}
}
}

\pgfkeys{/forest,
my rounded corners/.append style={rounded corners=2pt},
}
\begin{document}
%\title{A Survey on Image Based One Class Classification }
\title{One-Class Classification: A Survey}

\author{Pramuditha~Perera, \IEEEmembership{Member, IEEE}, Poojan Oza, \IEEEmembership{Student Member, IEEE}
	and~Vishal~M.~Patel, \IEEEmembership{Senior Member, IEEE}% <-this % stops a space
	\IEEEcompsocitemizethanks{\IEEEcompsocthanksitem P. Perera is with AWS AI - NYC, but the work is done at Johns Hopkins Univeristy. P. Oza and V. M. Patel are with the Department of Electrical and Computer Engineering at Johns Hopkins University, MD 21218.\protect\\
		
	}% <-this % stops an unwanted space
	\thanks{ }}

\maketitle

\maketitle

\begin{abstract}
One-Class Classification (OCC) is a special case of multi-class classification, where data observed during training is from a single positive class. The goal of OCC is to learn a representation and/or a classifier that enables recognition of positively labeled queries during inference. This topic has received considerable amount of interest in the computer vision, machine learning and biometrics communities in recent years. In this article, we provide a survey of classical statistical and recent deep learning-based OCC methods for visual recognition.   We discuss the merits and drawbacks of existing OCC approaches and identify promising avenues for research in this field.  In addition, we present a discussion of commonly used  datasets and evaluation metrics for OCC. 

%	In this survey, we discuss recent advents in OCC with an emphasis on computer vision applications. We survey both feature learning-based methods and classification-based methods in detail that target OCC. The survey concludes with a discussion of evaluation metrics, datasets and open problems in OCC.

%In this article, we provide a survey of OCC methods for visual recognition. We survey both feature learning-based methods and classification-based methods in detail that target OCC.  We discuss the merits and drawbacks of existing OCC approaches and identify promising avenues for research in this field.  In addition, we present a discussion of commonly used  datasets in OCC and metrics used for evaluating various OCC methods. 

\end{abstract}
%However, these systems often produce predictions with very high probability associated with fail when they encounter instances outside the set of training classes.  Deep-fool images and adversarial attack images are two classes of examples where this occurs.

\section{Introduction}
Over the last decade, methods based on Deep Convolutional Neural Networks (DCNNs) have shown impressive performance improvements for object detection and recognition problems.  Availability of large multi-class annotated datasets makes it possible for deep networks to learn discriminative features that a classifier can exploit to perform recognition. In this survey we focus on the extreme case of recognition -- One-Class Classification (OCC), where data from only a single class (labeled positive) is present during training. During inference, the classifier encounters data from both the positive class and outside the positive class (sometimes referred to as the negative class). The objective of OCC is  to determine whether a query object belongs to the class observed during training.  In the absence of multiple-class data, both learning features and defining classification boundaries become more challenging in OCC compared to multi-class classification. Nevertheless, it has applications in several image-based machine learning tasks. One-class classification methods are extensively used in  abnormal image detection \cite{Saleh:2013:OAD:2514950.2516141}, \cite{Bergmann_2019_CVPR} and abnormal event detection \cite{BMVC2017_139}, \cite{8484027}.  They are also used extensively  in biometrics applications such as Active Authentication \cite{DualMPM}, \cite{oza2019active} and anti-spoofing \cite{antispoofing}, \cite{as2}, \cite{Engelsma2019GeneralizingFS}, \cite{Yadav_2020_WACV}. Therefore,  there has been a significant interest in the computer vision,  machine learning and biometrics communities in developing OCC algorithms.

\begin{figure}
	\centering
	\includegraphics[width=1\linewidth]{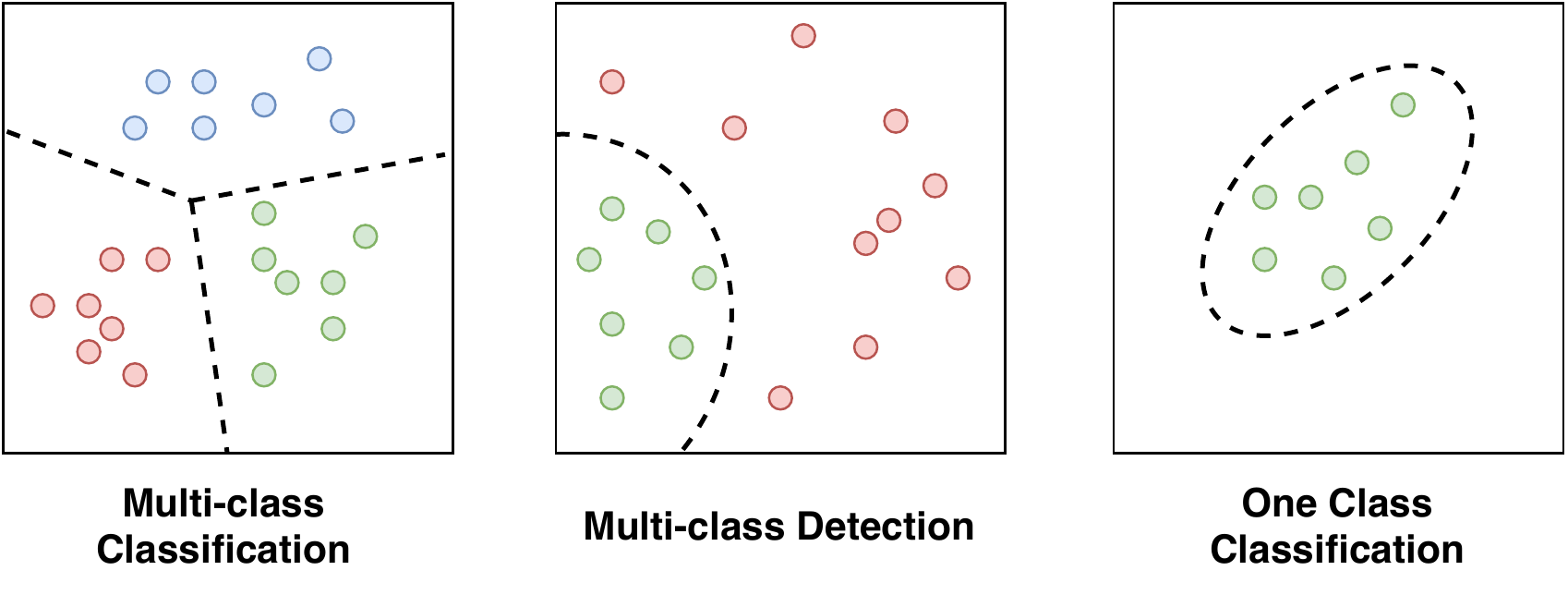}
	\vskip-10pt\caption{Different forms of classification. Data points corresponding to different classes are shown in different color. Both multi-class classification and detection contain data from different classes during training.  In OCC, only a single class is given to learn a representation and/or a classifier.}
	\label{fig:diff}
\end{figure}

In Figure~\ref{fig:diff} we illustrate the differences among OCC and multi-class detection and classification. In multi-class classification, training data from several classes are present. The classifier learns multiple decision boundaries to separate each class from the others. On the other hand, in detection, there exists data from two classes -- a positive class and a negative class. The learned classification boundary in this case separates out positive data from the negative data. In contrast, OCC has only data from the positive class during training. The learned classifier defines a boundary that encompasses positive data with the hope that it will provide good separation from the other objects in the world.

There exists multiple research areas in computer vision and machine learning that are closely related to the task of OCC. First, we discuss similarities and differences between OCC and these research topics.

\noindent \textbf{One-class novelty detection. } The objective of one-class novelty detection \cite{ocgan}, \cite{AND}, \cite{GPND} is the same as of OCC. Here, a detector is  sought that can detect objects from the observed class. Learned detector can be transformed into a classifier by selecting a suitable detection threshold.  Therefore, OCC and a one-class novelty detection solve essentially the same problem.  In this survey, we make no distinction between OCC and one class novelty detection.%The only difference between the two problems is how performance is evaluated. It is customary for performance to be evaluated using classification accuracy or F1 score in the case of OCC.  On the other hand, area under the ROC curve is used commonly when performance is evaluated in one-class novelty detection. We make no distinction between the two.

\noindent \textbf{Outlier detection (unsupervised anomaly detection). }  In outlier detection \cite{jhu}, \cite{Lai2020Robust}, \cite{outlier}, a mixture of normal and abnormal data is presented without ground truth annotations. The objective is to separate normal data from abnormal data using unsupervised techniques. In contrast, all training data is assumed to be normal in OCC.  Therefore, OCC is considered to be a supervised learning problem whereas outlier detection is unsupervised.  One class classification solutions cannot be directly applied to outlier detection problems and vise versa.

\noindent \textbf{Open-set recognition. }  Open-set recognition \cite{Scheirer_2014_TPAMIb}, \cite{BendaleB16}, \cite{SROSR}, \cite{Oza_2019_CVPR}, \cite{ECCV18}, \cite{gopenmax}, \cite{Perera_2019_CVPR} is an extension to multi-class classification. Given a query image, open-set recognition considers the possibility of the query not belonging to any of the classes observed during training. Therefore, open-set recognition essentially learns $C+1$ decision boundaries for a $C$-class problem, where the additional boundary separates the known classes from the novel (open-set) classes. One class classification is the extreme case of open-set recognition where $C=1$.

Several surveys exist in the literature on OCC and related techniques \cite{Chandola:2009:ADS:1541880.1541882}, \cite{occ_survey}, \cite{pang2020deep}, \cite{pimentel2014review}, \cite{chalapathy2019learning}, \cite{khan2014one}, \cite{khan2009survey}.  However, some of them present generic one class techniques that are designed for low dimensional inputs and do not always generalize well to image data \cite{Chandola:2009:ADS:1541880.1541882}, \cite{occ_survey}. Reviews provided in \cite{pang2020deep}, \cite{pimentel2014review}, \cite{chalapathy2019learning} only focus on the specific applications of OCC such as image-based novelty and anomaly detection. Furthermore, survey papers \cite{khan2014one}, \cite{khan2009survey} do not include OCC methods proposed over the last few years, especially deep learning-based methods.  Whereas in this paper, we present a survey of recent advances in OCC with special emphasis on deep features and classifiers.  

This paper is organized as follows. In Section~\ref{sec:taxonomy}, we present a taxonomy for OCC. In Section~\ref{sec:features}, we discuss feature learning techniques targeting OCC. In Section~\ref{sec:OCC_Alg}, we first present a review of classical OCC algorithms and then discuss recent  deep learning-based algorithms in detail.  Section~\ref{sec:data_and_metrics} presents a discussion of commonly used datasets and evaluation metrics for OCC. Finally, Section~\ref{sec:openproblems} concludes the paper with a brief summary and discussion.

\section{Taxonomy}\label{sec:taxonomy}
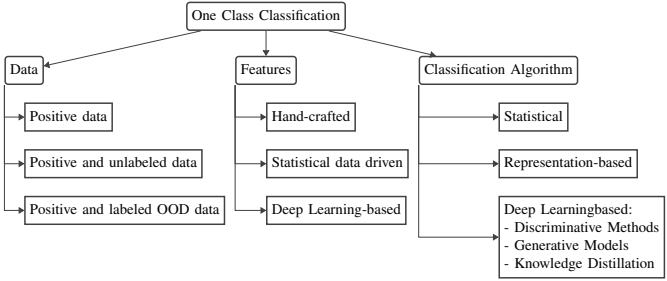
\begin{figure}
	\centering
	\resizebox{1\linewidth}{!}{
		\begin{forest}
			for tree={
				line width=1pt,
				draw=linecol,
				%     drop shadow,
				fit=rectangle,
				edge={color=linecol, >={Triangle[]}, ->},
				where level=0{%
					l sep+=5pt,
					calign=child,
					calign child=2,
					%  inner color=green!80,
					%   outer color=green,
					align=center,
					my rounded corners,
					for descendants={%
						calign=first,
					},
				}{%
					where level=1{%
						%inner color=green!80,
						%outer color=green,
						my rounded corners,
						align=center,
						parent anchor=south west,
						tier=three ways,
						for descendants={%
							child anchor=west,
							parent anchor=west,
							align=left,
							anchor=west,
							%            inner color=MistyRose1!80,
							%           outer color=MistyRose1,
							edge path={
								\noexpand\path[\forestoption{edge}]
								(!to tier=three ways.parent anchor) |-
								(.child anchor)\forestoption{edge label};
							},
						},
					}{}%
				},
			}
			[One Class Classification
			[Data
			[Positive data
			[Positive and unlabeled data
			[Positive and labeled OOD data]
			]
			]
			]
			[Features 
			[Hand-crafted
			[Statistical data driven
			[Deep Learning-based 
			]
			]
			]
			]
			[Classification Algorithm
			[Statistical
			[Representation-based 
			[Deep Learningbased:\\
			 - Discriminative Methods\\
			 - Generative Models\\
     - Knowledge Distillation
			]
			]
			]
			]
			]
	\end{forest}}
	\caption{A taxonomy for one class classification methods. These methods can be broadly categorized by the type of training data used, feature representations used and learning method used for modeling one-class data. \label{fig:tax}}
\end{figure}

Considering the research work carried out in computer vision and machine learning communities, we propose a taxonomy for the study of image-based OCC problems as shown in Figure~\ref{fig:tax}.  Main categories identified in the taxonomy are as follows:

\begin{enumerate}
	\item \textbf{Data.} Learning with positive data, positive and unlabeled data and learning with positive data in the presence of labeled Out Of Distribution (OOD) data.
	\item \textbf{Features.} Handcrafted features, statistical data driven features and deep-learning based features.
	\item \textbf{Classification Algorithm.} Statistical classification methods, representation-based classification methods and deep-learning based methods.
\end{enumerate}

The taxonomy categorizes OCC methods based on training data usage, feature and classification method used. Contributions of each work described falls under one or more aforementioned categories.  In this paper, we discuss features and classifiers used in the recent literature for OCC. Unless otherwise specified, all methods are designed to train with only positively labeled data belonging to the same class.

%	\begin{comment}

%	\end{comment}

\begin{figure*}
	\centering
	\includegraphics[width=1\linewidth]{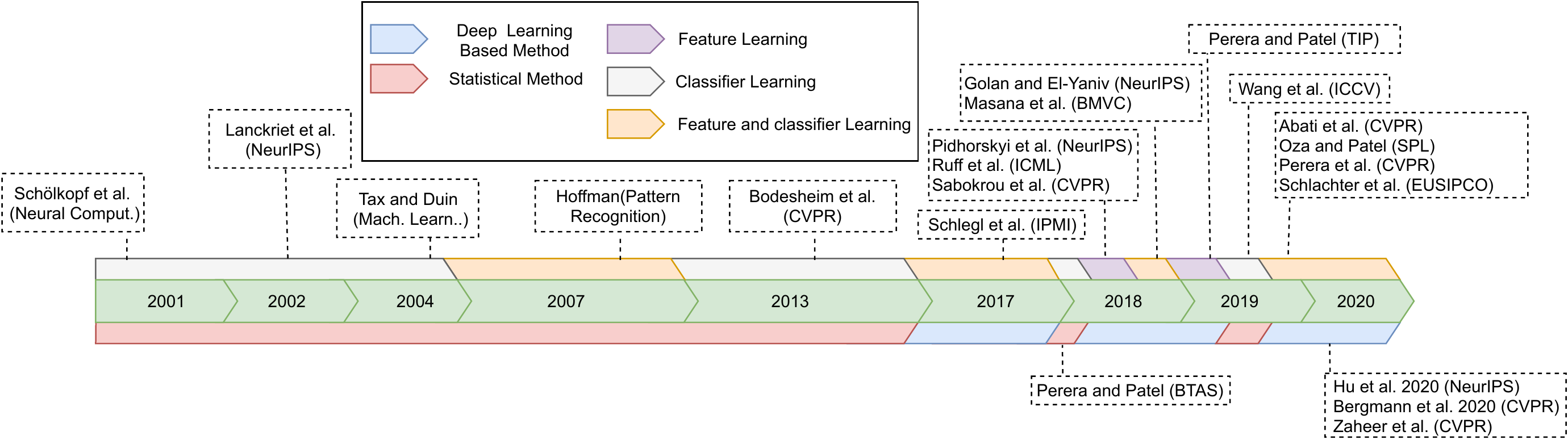}
	\vskip -2.5pt
	\caption{Landmarks of one-class classification showing the evolution of methods over the years. As we can see the recent trend in one-class classification largely focuses on developing deep learning-based methods.}
	\label{fig:time}
\end{figure*}

In Figure~\ref{fig:time},  landmarks of OCC are illustrated with a break down of their contributions (feature learning, classifier learning, both feature and classifier learning).  We further indicate whether each method is based on a statistical learning framework or a deep learning framework in Figure~\ref{fig:time}. Initial works on OCC primarily used statistical features and focused on developing classifiers.  Most methods since 2017 have used deep features in their frameworks. These methods either use classical classifiers on deep features or simultaneously learn both features and classifier. In Table~\ref{tbl:sum} we summarize papers we survey in this work and specify their contributions with respect to the taxonomy we provided.

\begin{table*}[] \label{tbl:sum}
	\centering
			\caption{Summary of works surveyed in this paper. This table indicates whether the contribution of each work is in the classifier or in the feature (or both).}
	\begin{tabular}{|l|l|l|l|l|}
		\hline
		\textbf{Method}               & \textbf{Features} & \textbf{Classifier} & \textbf{Data}  & \textbf{Publication}                      \\ \hline\hline
		Kernel PCA                    & Statistical       & Representation      & Positive       & Hoffman  2007 (Pattern Recognition) \cite{HOFFMANN2007863}  \\ \hline
		Geometric Transformations      & Deep              & -                   & Positive       & Golan and El-Yaniv 2018 (NeurIPS) \cite{rotations}              \\ \hline
		Deep Metric Learning          & Deep              & -                   & Positive + OOD & Masana et al. 2018 (BMVC)   \cite{masana2018metric}              \\ \hline
		Feature Learning With OOD Data (DOC) & Deep              & -                   & Positive + OOD & Perera and Patel 2019 (TIP)     \cite{ocfeatures}             \\ \hline
		One-class SVM                          & -                 & Statistical         & Positive       & Schölkopf et al. 2001 (Neural Comput.)  \cite{Scholkopf:2001:ESH:1119748.1119749}  \\ \hline
		SVDD                          & -                 & Statistical         & Positive       & Tax and Duin 2004 (Mach. Learn.)    \cite{Tax:2004:SVD:960091.960109}       \\ \hline
		OCMPM                         & -                 & Statistical         & Positive       & Lanckriet et al. 2002 (NeurIPS)      \cite{lank}     \\ \hline
		DS-- OCMPM                    & -                 & Statistical         & Positive       & Perera and Patel 2018 (BTAS)    \cite{DualMPM}             \\ \hline
	KNFST                       & -                 & Representation      & Positive       & Bodesheim et al. 2013 (CVPR)   \cite{Bodesheim_2013_CVPR}            \\ \hline
		
		GODS                         & -           & Statistical                & Positive       & Wang et al. 2019 (ICCV)   \cite{Wang_2019_ICCV}             \\ \hline
		OCCNN                         & Deep              & Deep                & Positive + OOD      & Oza and Patel 2019 (SigPro Letters)   \cite{oza2019one}       \\ \hline
		Deep SVDD                     & Deep              & Deep                & Positive       & Ruff et al. 2018 (ICML)         \cite{dsvdd}          \\ \hline
		AnoGAN                        & Deep              & Representation      & Positive       & Schlegl et al. 2017 (IPMI)       \cite{IPMI}         \\ \hline
		ALOCC                         & Deep              & Deep     & Positive       & Sabokrou et al. 2018 (CVPR)      \cite{cvpr2018}         \\ \hline
		OCGAN                         & Deep              & Representation      & Positive       & Perera et al. 2019 (CVPR)     \cite{ocgan}            \\ \hline
		PGND                          & Deep              & Deep                & Positive       & Pidhorskyi et a. 2018 (NeurIPS)    \cite{GPND}       \\ \hline 
		AND                           & Deep              & Deep                & Positive       & Abati et al. 2019 (CVPR)    \cite{AND}              \\ \hline
		ICS                           & Deep              & Deep                & Positive       & Schlachter et al. 2019 (EUSIPCO)    \cite{schlachter2019}             \\ \hline
		P-KDGAN                   & Deep              & Deep                & Positive       & Zhang et al. 2020 (IJCAI) \cite{zhangp}   \cite{AND}              \\ \hline
		HRN          & Deep              & Deep                & Positive       & Hu et al. 2020 (NeurIPS) \cite{hu2020hrn}          \\ \hline
		US-OCL                   & Deep              & Deep                & Positive       & Bergmann et al. 2020 (CVPR) \cite{bergmann2020uninformed} \\ \hline
		OGN                   & Deep              & Deep                & Positive       & Zaheer et al. 2020 (CVPR) \cite{zaheer}             \\ \hline
		
	\end{tabular}	
\end{table*}

%%%%%%%%%%%%%%%%%%%%%%%%%%%%%%In Table~\ref{tbl:not}, notations used in the paper are defined along with their definitions.

\section{Features for OCC}\label{sec:features}
Learning features that aid in classification is a well researched problem in multi-class classification. In order to perform classification correctly,  a suitable feature should be selected such that a classifier can define regions where objects from each class appear consistently. Similar to multi-class classification, OCC will benefit from features that position data of the positive class separately from the other objects in the world. However, learning/selecting such a feature is a more difficult task as there are no non-positive data available during training. Nevertheless, following two properties can be identified in a desired feature for one class classification \cite{ocfeatures}:

\begin{enumerate}
	\item  \textbf{Compactness.} A desired quality of a feature is to have a similar feature representation for different images of the same class. Hence, a collection of features extracted from a set of images of a given class will be compactly placed in the feature space. 
	\item \textbf{Descriptiveness.} The given feature should produce distinct representations for images of different classes. Ideally, each class will have a distinct feature representation from each other.
\end{enumerate}

Earliest of features in computer vision were engineered manually with the objective of providing good description of the image \cite{hog}, \cite{lbp_face}, \cite{Lowe04distinctiveimage}. These features are known as hand-crafted features. Later, researchers used data-driven approaches to discover more discriminative features through optimization. In\cite{vapnik95}, intra-class distance was minimized while maximizing inter-class distance to learn an informative feature embedding. It was later shown by \cite{Wright:2009:RFR:1495801.1496037} that sparse representation based on a dictionary can lead to better discriminative features. More recent works on feature learning use deep-learning with cross-entropy loss and triplet similarity loss to learn features for multi-class classification.

Traditional hand-crafted features  \cite{hog}, \cite{lbp_face} can be used for OCC as well. Use of hand crafted features are not any different from multiple-class classification in this case. On the other hand, data-driven feature learning methods such as LDA \cite{vapnik95} and deep-feature learning do not fit well in OCC framework due to the absence of multiple classes. However, data-driven methods that do not use class label information for feature learning such as PCA \cite{Turk1991}, KPCA \cite{kpca_theory} and sparse coding can be used for one class feature learning. Usage of these data-driven methods for one class feature extraction is not different from the methodology used in multi-class classification. 

\subsection{Statistical Features} \label{lbl:stat_features}

\noindent \textbf{Sparse Coding.} Given a query image, sparse representation-based methods extract features with respect to a dictionary \cite{Wright:2009:RFR:1495801.1496037}. In the simplest case, the dictionary can be constructed by stacking training data in columns of the data matrix $\mathbf{X}=[\mathbf{x}_1, \mathbf{x}_2,\cdots, \mathbf{x}_N]\in\mathbb{R}^{n \times N}$, where have assumed that we have a dataset with $N$ number of training images, $\mathcal{D}=\{x_i\}_{i=1}^{N}$, and $x_i \in \mathbb{R}^{3 \times H \times W}$.  Note that $\mathbf{x}_i \in \mathbb{R}^n$, corresponds to the vectorized version of the $i$th image, where $n = 3 \cdot H \cdot W$. Each column of the dictionary is referred to as an atom. For a given vectorized test image $\mathbf{x}_{test} \in \mathbb{R}^n$, the following sparsity promoting optimization problem is solved as:
%a sparse set of weights are calculated that allows to approximately represent $\mathbf{x}_{test}$ as a linear combination of atoms in $\mathbf{X} \in \mathbb{R}^{N \times n}$. This is done by solving the following optimization problem:
\begin{equation}\label{eq:src}
\hat{\mathbf{v}}_{test} \ = \  \arg \min_{\mathbf{v}_{test}}  \|\mathbf{v}_{test} \|_1  ~\;\;\text{s.t.}\;\; ~ \mathbf{X} \mathbf{v}_{test} = \mathbf{x}_{test},
\end{equation}
where $\|\mathbf{v}_{test}\|_1$ denotes the $\ell_{1}$-norm of $\mathbf{v}_{test}$.
By solving Eq.~\ref{eq:src}, we get a sparse code $\hat{\mathbf{v}}_{test}$ which can be used as a feature for one-class classification. In \cite{song2016one}, Song \emph{et al.} utilized sparse representations to perform OCC for a remote sensing application.

Instead of using a pre-determined dictionary, one can directly learn a dictionary from the data which can lead to a more compact representation and hence better classification \cite{Olshausen_Field}. Algorithms such as method of optimal directions (MOD) \cite{MOD}, K-SVD \cite{KSVD}, \cite{Kernel_KSVD}, and convolutional sparse coding (CSC) \cite{CSC} can be used to learn a dictionary from training data.\\

\noindent \textbf{PCA and Kernel PCA.} 
%Let us consider a dataset with $N$ number of training images, $\mathcal{D}=\{x_i\}_{i=1}^{N}$, where each image $x_i \in \mathbb{R}^{3 \times H \times W}$. Furthermore, let us create a data matrix $\mathbf{X} \in \mathbb{R}^{N \times n}$ by stacking vectorized images as columns, i.e., $\mathbf{X} = [\mathbf{x}_1, \mathbf{x}_2, \dots, \mathbf{x}_N]$, where $\mathbf{x}_i  \in \mathbb{R}^n$ is $i^{th}$ vectorized image and $n = 3 \cdot H \cdot W$. 
Principle Component Analysis (PCA) finds a lower dimensional subspace that best accounts for the distribution of images in the image space \cite{Turk1991}. This subspace is shown to be spanned by eigenvectors of the covariance matrix of the dataset. Firstly, the covariance matrix $\mathbf{C}$  of the data is calculated as in $\mathbf{C} = \frac{1}{N} \sum_{i=1}^{N} (\mathbf{x}_i-\bm{\mu}) (\mathbf{x}_i- \bm{\mu})^T$, where $\bm{\mu}$ is the mean calculated as $\bm{\mu} = \frac{1}{N} \sum_{i=1}^{N}\mathbf{x}_i$.  In PCA, a set of eigenvectors $\{\mathbf{v}_1, \mathbf{v}_2, \dots, \mathbf{v}_N\}$ corresponding to the eigenvectors of $\mathbf{C}$ are found. Given a vectorized image $\mathbf{x}_{test}$, it can be projected on to the eigenspace defined by vectors $\{\mathbf{v}_1, \mathbf{v}_2, \dots, \mathbf{v}_N\}$, where the magnitude projection on the $v_i^{th}$ eigenvector $\|\mathbf{v}_i\|_2$ is given by $\mathbf{v}_i^T(\mathbf{x}_{test}-\bm{\mu})$.  Then, a feature is defined by considering the collection  of magnitudes $ \mathbf{v}_{test} = [\|\mathbf{v}_1\|_2, \|\mathbf{v}_2\|_2, \dots, \|\mathbf{v}_N\|_2]$.

Kernel PCA extends PCA to handle the non-linearity of data  \cite{kpca_theory}. Let us assume that we have a mapping function $\Phi : \mathbb{R}^n \rightarrow \mathbb{R}^d$ that can map data $\mathbf{x}_i$ into a feature space $\Phi(\mathbf{x}_i)$. Depending on the mapping function, the feature space dimension $d$ can be arbitrarily large. For simplicity let us assume the data mapped into the feature space is centered, i.e., $\frac{1}{N} \sum_{i=1}^{N}\Phi(\mathbf{x}_i)=0$. As a result, the covariance matrix in the feature space can be computed as follows:
\begin{equation}
	\bar{\mathbf{C}} = \frac{1}{N}\sum_{i=1}^{N}\Phi(\mathbf{x}_i)\Phi(\mathbf{x}_i)^T,
\end{equation}
To perform PCA in the feature space, we need to find a set of eigenvectors and non-zero eigenvalues in the feature space. Let $\lambda_j$ and $\mathbf{v}_j$ denote the eigen values and the corresponding eigen vector of $\bar{\mathbf{C}}$, respectively.  
One can represent eigen vectors as $\mathbf{v}_j = \sum_{i=1}^{N} \alpha^j_i \Phi(\mathbf{x}_i)$, where  $\alpha^j_1, \alpha^j_2, \dots, \alpha^j_N$ are a set of coefficients. Since the eigenvalue and eigenvectors must satisfy $\lambda_j \mathbf{v}_j = \bar{\mathbf{C}} \mathbf{v}_j$, for performing PCA in the feature space, we consider an equivalent system as:
%\begin{equation}
%	\lambda (\Phi(\mathbf{x}_i) \cdot \mathbf{V}) = (\Phi(\mathbf{x}_i) \cdot \bar{\mathbf{C}}\mathbf{V}), \ \  \text{for} \ i=1, 2, \dots, N
%\end{equation}
\begin{flalign}
\lambda_j \langle \Phi(\mathbf{x}_i), \mathbf{v}_j \rangle = \langle \Phi(\mathbf{x}_i), &\bar{\mathbf{C}}\mathbf{v}_j \rangle, \ \  \text{for} \ i=1,\dots, N,\\
\lambda_j \langle \Phi(\mathbf{x}_i), \sum_{l=1}^{N} \alpha^j_i \Phi(\mathbf{x}_l) \rangle &= \langle \Phi(\mathbf{x}_i), \bar{\mathbf{C}} \sum_{l=1}^{N} \alpha^j_i \Phi(\mathbf{x}_l) \rangle
\label{eq:kpca_1},
\end{flalign}
where, $\langle \cdot, \cdot \rangle$ indicates inner product between two vectors. Let us define a column vector as $\bm{\alpha}^j = [\alpha^j_1, \alpha^j_2, \dots, \alpha^j_N]^{T}$ and a non-singular $N \times N$ matrix $\mathbf{K}$, where each entry of the matrix is defined as, $\mathbf{K}_{rt} := \langle \Phi(\mathbf{x}_r), \Phi(\mathbf{x}_t)  \rangle$.  Using these notations, we can simplify the above equation as:
\begin{equation}\label{eq:kpca_2}
	N \lambda_j \bm{\alpha}^j =  \mathbf{K} \bm{\alpha}^j.
\end{equation}
Then the solution $\bm{\alpha}^j$ belonging to non-zero eigenvalues is found by requiring that corresponding vectors in $\mathbb{R}^d$ be normalized, i.e., $\langle \mathbf{v}_j, \mathbf{v}_j \rangle = 1$ \cite{welling2005kernel}. Consequently, we can find the corresponding eigenvectors $\mathbf{v}_j$. For a new vectorized test image $\mathbf{x}_{test}$, we can compute its projection onto the eigenvectors. We can calculate the projection with respect to the $j^{th}$ eigenvector in the feature space as:
\begin{equation}\label{eq:kpca_3}
\langle \mathbf{v}_j, \Phi(\mathbf{x}_{test}) \rangle = \sum_{i=1}^{N} \alpha^j_i \langle \Phi(\mathbf{x}_i), \Phi(\mathbf{x}_{test}) \rangle.
\end{equation}
Note that in Eq.~\ref{eq:kpca_3}, we do not require the definition of mapping function $\Phi$ in its explicit form and only need it in the dot product form. Hence, we can define functions to perform dot products without actually having to explicitly map data points using the mapping function $\Phi$. These functions are commonly referred to as kernel functions \cite{hempstalk2008one}. An example of a kernel function is the the RBF kernel \cite{hempstalk2008one}, defined as $K(\mathbf{x}_i, \mathbf{x}_j) = e^{-\|\mathbf{x}_i-\mathbf{x}_j\|^2/{2\sigma^2}},$ where $\sigma$ is a parameter of the kernel.  Similar to PCA, the projection values computed using Eq.~\ref{eq:kpca_3} are used as a feature representation for any test data. Note that, kernel PCA can be derived for data without the assumption that data is centered in feature space with minor modification in the derivation process \cite{scholkopf1997prior}. 

Both PCA and kernel PCA produce features that are descriptive as the lower dimensional subspace learned has the capacity to represent the image space of training data \cite{HOFFMANN2007863}. However, the learned embeddings are not necessarily compact.

\subsection{Deep learning-based Features}
\begin{figure}
	\centering
	\includegraphics[width=0.7\linewidth]{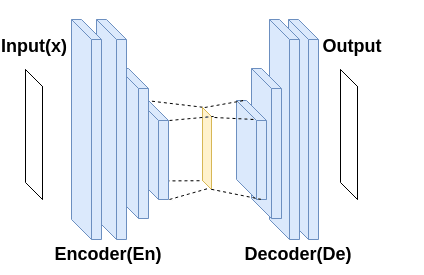}
	\vskip -2.5pt
	\caption{Learning features with the help of a stacked auto-encoder network trained to reconstruct the input images using mean-squared error.}
	\label{fig:ae}
\end{figure}

\noindent \textbf{Deep Auto-encoders}\\	
An auto-encoder is a (convolutional) neural network consisting of an encoder ($\mbox{En}$) and decoder ($\mbox{De}$) networks as shown in  Figure~\ref{fig:ae}.  The encoder comprises of a set of convolution layers followed by activation functions. The decoder consists of transposed convolutional layers and commonly has a structure similar to that of an inverted encoder.  An auto-encoder is trained with the objective of minimizing the distance between the input and the output of the network. In theory, any distance measure, such as:
\begin{equation}
\mathcal{L}_{mse} = \|{x} - \mbox{De}(\mbox{En}({x})) \|_2^2,
\end{equation}
can be considered to learn the parameters of the auto-encoder, where ${x}$ is the input image. 

It is the usual practice to have a bottleneck latent-space in between with a dimension smaller than the input. Due to this bottleneck, auto-encoder retains only the essential information in the latent space which is required for reconstruction. These encoder features are informative and can be expected to exhibit \textit{descriptiveness}. It has been shown in the literature that adding noise to the input can further improve the quality of the learned representation by reducing the problem of over-fitting, making it more generalizable \cite{Vincent:2008:AE}.  When noise is added to the input, the network is referred to as a \textit{de-noising auto-encoder} \cite{Vincent:2008:AE}. In a de-noising auto-encoder, given a noisy image, the network is expected to reconstruct the clean version of the image.\\
%De-noising auto-encoders open up the possibility of having a latent dimension larger than the input image dimension \cite{Vincent:2008:AE}. \\

\noindent \textbf{Geometric Transformation based Self-supervision}\\
Self supervision is a machine learning technique that can be used to learn informative representations from unlabeled data. Golan \emph{et al.} showed that self supervision can yield representations that work in favor of one class classification \cite{Gidaris2018UnsupervisedRL}, \cite{rotations}. In their work, a random geometric transform chosen from a pre-determined set of transforms is first applied to every input image during training. The network is trained to correctly predict the applied transformation. Let us denote a set of $k$ geometric transformations as $\{\mathcal{T}_1 ,\mathcal{T}_1 ,\dots,\mathcal{T}_{k}  \}$, where for any given image ${x}$ both ${x}, \mathcal{T}_i({x}) \in \mathbb{R}^{H \times W}$ and $\mathcal{T}_1$ denotes identity transformation. During training, for a given input, a number $0 \geq r \geq k$ is randomly chosen. The input image ${x}$ is transformed using the $r^{th}$ transformation to arrive at $\mathcal{T}_r({x})$. The transformed image is passed through a convolutional neural network and the network parameters are optimized with a cross-entropy loss where $r$ is considered to be the ground truth label. The trained network produces features suitable for one class classification.

Golan \emph{et al.} \cite{rotations} showed that a neural network trained in this manner produces relatively higher probability scores in the ground truth class for samples of the positive class. For a given test image ${x}_{test}$, \cite{rotations} proposed to evaluate a normality score $s_{test}$ by considering log likelihood probability of the ground truth class with all $k$ transformations as:
\begin{equation}
s_{test}({x}_{test}) = \sum_{i=1}^{k} \log(F(\mathcal{T}_i({x}_{test}))),%| \mathcal{T}_i
\end{equation}
where $F(\cdot)$ is the softmax output of the network. Furthermore, Golan \textit{et al.} \cite{rotations} derived a Dirichlet distribution-based score which is more effective for one class classification. This second score was derived by assuming that each conditional distribution  takes the form of a Dirichlet distribution  $F(\mathcal{T}_i({x}))|\mathcal{T}_i \sim Dir(\mathbf{a}_i)$ where $\mathbf{a}_i \in \mathbb{R}_{+}^{k}$. When estimates of Dirichlet parameters $\hat{\mathbf{a}}_i$ are found with the help of maximum likelihood estimation, the final normality score denoted as $u_{test}$ can be expressed as: 
%\begin{equation}
%\begin{multlined}
%n_s(\mathbf{x}) = \sum_{i=0}^{k-1} [  \log \Gamma ( \sum_{j=0}^{k-1}[\hat{\alpha_i}]_j ) -   \sum_{j=0}^{k-1} \log \Gamma  ([\hat{\alpha_i}]_j ) \\ +   \sum_{j=0}^{k-1}([\hat{\alpha_i}]_j -1) \log f( \mathcal{T}_i(\mathbf{x}) )_j ].
%\end{multlined}
%\end{equation}
%When constant terms with respect to the input $x$ are ignored, the Dirichlet normality score simplifies to,
\begin{equation}
	u_{test}(x_{test}) = \sum_{i=1}^{k} (\hat{\mathbf{a}}_i - 1) \cdot f(\mathcal{T}_i({x})),
\end{equation}

\noindent \textbf{Deep Metric Learning with OOD Data} \\
A contrastive loss-based metric learning method is proposed in \cite{masana2018metric} to learn features for OCC.  For the metric learning process, \cite{masana2018metric}  proposes to use data from an OOD dataset. In the absence of an OOD dataset, it is artificially generated by adding random Gaussian noise to the images. Let $F$ be the network function of a deep convolutional network. For a pair of input images ${x_1}$ and ${x_1}$, the distance between the to inputs in the feature space is defined as $K(x_1, x_2) = \| F( {x_1}) - F( {x_2})\|_2$.  The training process incorporates two types of labels. Label $\gamma$ indicates whether the two inputs belong to the same class $(\gamma=0)$ or not $(\gamma=1)$. Label $\zeta$ denotes if both images belong to a OOD dataset $(\zeta = 0$) or not ($\zeta = 1$). The contrastive loss is defined as:
\begin{equation}
\begin{array}{llllll}
\mathcal{L} = &\frac{1}{2} (1 - \gamma) \cdot \zeta \cdot K(x_1, x_2)^2 \\
& \ \ \ \ \ \ \ \  \ \ + \frac{1}{2} \gamma \cdot \zeta \cdot (\max (0, m-K(x_1, x_2))^2,
\end{array}
\end{equation}
where, $m$ denotes minimum margin. For positively labeled data when both images are from the same class ($\gamma=0, \zeta = 1$), the loss becomes  $\frac{1}{2} K(x_1, x_2)^2$. Therefore, during training network learns a similar embedding for positive data. As a result, the compactness property is satisfied. On the other hand, when the image pair is from different classes, the embedding is encouraged to be at least $m$ distance apart in the feature space. In the case when both images are from the same class ($\gamma=0, \zeta = 0$), the loss becomes zero. Therefore, the learned feature embedding becomes descriptive with the ability of differentiating positively labeled and data outside the given category.\\

\noindent \textbf{Feature Learning With OOD Data (DOC)}\\
In \cite{ocfeatures}, authors consider the special scenario where a set of labeled Out Of Distribution (OOD) data is available during training along side positive data. These OOD data are data collected from non-overlapping problem domain. For example, in the case of a face recognition problem for one-class classification an annotated object dataset can be considered as OOD. This setting is important in practice as many real life applications of OCC can be solved in this setting. Let us consider a deep network with a feature extraction sub-network $F$ and a classification sub-network $G$. The network $G \circ F $ is first trained using the OOD data using cross entropy loss. Let us denote the positively labeled dataset with $N$ number of training images as $\mathcal{D}_t=\{x_{ti}\}_{i=1}^{N}$ and OOD dataset with $M$ number of images as $\mathcal{D}_r=\{x_{ri}, y_{ri}\}_{i=1}^{M}$, where images $x_{ti}, x_{ri} \in \mathbb{R}^{3 \times H \times W}$ and $y_{ri}$ is the target label for the image $x_{ri}$. The features extracted from any image $x_{ti}$ are denoted as $F(x_{ti}) = \mathbf{z}_{ti} \in \mathbb{R}^d$. Here, $d$ is the dimension of the feature space. A \textit{compactness loss $\mathcal{L}_{c}$} is defined that measures compactness of the learned feature with respect to positive data. The \textit{compactness loss} is evaluated using a intra-class distance for that given batch in the feature space as:
% Consider a data batch sampled from the dataset $\mathcal{D}$ having $B \ (<N)$ number of images. 
\begin{equation}\label{eq:lb}
\mathcal{L}_{c}(x_{ti}) = \frac{1}{d} (\mathbf{z}_{ti} - \bm{\mu}_{ti})^T(\mathbf{z}_{ti}- \bm{\mu}_{ti}),
\end{equation}
where feature extracted from each image $x_{ti}$ $\mathbf{z}_{ti} \in \mathbb{R}^{d}$. The mean vector $\bm{\mu}_{ti} $ is defined as $\bm{\mu}_{ti}  = \frac{1}{N-1}\sum_{j \neq i} \mathbf{z}_{tj}$. Additionally, a \textit{descriptiveness loss} denoted as $\mathcal{L}_{d}$ is used to measure descriptiveness of the learned features. It is measured using cross-entropy loss  obtained by the output of the network with respect to OOD data. The network is fine-tuned by jointly optimizing over both \textit{compactness} and \textit{descriptiveness} loss as:
\begin{equation}\label{eqn:opt2}
\min_{F, G} ~ \sum_{i=1}^{M} \mathcal{L}_{d}({x}_{ri}, y_{ri}) + \lambda  \sum_{i=1}^{N} \mathcal{L}_{c}({x}_{ti}),
\end{equation}
where, $x_{ri}$ and $x_{ti}$ denote data from the OOD dataset and the positively labeled data respectively and $\lambda$ is a hyper-parameter. As a result, the learned feature is both descriptive and compact with respect to the positively labeled data.

\section{OCC Algorithms}\label{sec:OCC_Alg}

\subsection{Representation-based Methods}
In Section.~\ref{lbl:stat_features}, we discussed statistical methods that can be used to obtain features for one class classification. These methods encode information present in the input image to a feature vector. Once the feature corresponding to any query image is obtained, it can be used to perform classification using two different strategies. In the first strategy, the feature is mapped back to the image space to obtain a reconstructed image. Then, reconstruction error is calculated in the image space. In the second strategy, the feature representation is used to calculate a distance measure which can encode the score of query image with respect to training data. The sparse coding and PCA falls into the former strategy. On the other hand, kernel based one class classification utilizes the latter strategy. The detailed discussion on sparse coding and PCA/Kernel PCA based one-class methods is provided in Section.~\ref{lbl:stat_features}. Here, we focus on another method that obtains a score measure for any query image using Foley-Sammon Transform (FST) to learn a ``null space'' of the training data.\\

\noindent \textbf{Kernel Null Foley--Sammon Transform (KNFST)}\\
FST or Fisher transform is commonly used for linear discriminant analysis in the field of subspace methods \cite{foley1975optimal}. Null Foley--Sammon Transform (NFST) \cite{Bodesheim_2013_CVPR} is a special case of FST or Fisher transform, which is used to learn transformation from multi-class data. 
%Let us consider a dataset with $N$ number of training images, $\mathcal{D}=\{x_i\}_{i=1}^{N}$, where each image $x_i \in \mathbb{R}^{3 \times H \times W}$. Furthermore, let us denote vectorized images as, $\{\mathbf{x}_1, \mathbf{x}_2, \dots, \mathbf{x}_N\}$, where $\mathbf{x}_i  \in \mathbb{R}^n$ is $i^{th}$ vectorized image and $n = 3 \cdot H \cdot W$. 
The goal of NFST is to learn a mapping $\mathbf{w}$, where intra-class distances are zero while inter-class distances remain positive by solving the following equations for $\varphi$:
\begin{equation} \label{eq:sw}
\mathbf{w}^T\mathbf{C}_w\mathbf{w} = 0,
\end{equation} 
\begin{equation}
\mathbf{w}^T\mathbf{C}_b\mathbf{w} > 0,
\end{equation}
where, $\mathbf{C}_b$ and $\mathbf{C}_w$ are the inter-class/between-class and intra-class/within-class covariance/scatter matrix, respectively. When the sample size is small $n \leq d$, there exists $k-1$ null projection directions $ \mathbf{w}^{(1)}, \dots, \mathbf{w}^{(k-1)} $, for a $k$ class problem.  When the total scatter is defined as $\mathbf{C}_t = \mathbf{C}_b+\mathbf{C}_w$ and when $\mathbf{w}^T \mathbf{C}_w\mathbf{w} = 0$  holds, the objective boils down to finding $\mathbf{w}$ with  $\mathbf{w}^T\mathbf{C}_b\mathbf{w}>0$.

It was shown in \cite{Bodesheim_2013_CVPR} that these two conditions are satisfied when $\mathbf{w}^{(1)}, \dots, \mathbf{w}^{(k-1)}  \in ({Z}_t^{\perp} \cap {Z}_w)$ where $ {Z}_w$ is the null space of $\mathbf{C}_w$ and ${Z}_t^{\perp} $ is the row space of $ \mathbf{C}_t$. We can show that $Z_t^{\perp}$ can be exactly spanned with $\mathbf{x}_1 - \bm{\mu}, \dots, \mathbf{x}_N - \bm{\mu}$ with $\bm{\mu} = \frac{1}{N} \sum_{i=1}^{N}\mathbf{x}_N$. Therefore, to ensure $\mathbf{w} \in Z_t^{\perp}$ we can represent each $\mathbf{w}$ with weighted combination of some orthonormal basis $\{\mathbf{b}_1, \dots, \mathbf{b}_n\}$ as:
\begin{equation}
\mathbf{w} = \beta_1\mathbf{b}_1 + \dots + \beta_n\mathbf{b}_n,
\end{equation}
where, $\beta_i$ are scalar coefficients, $\mathbf{b}_i \in \mathbb{R}^{n}$ and $n \leq N$. The basis $\{\mathbf{b}_i\}_{i=1}^n$ can be obtain through Gram-Schmidt orthonormalization or principle component analysis. Let us denote a vector $\bm{\beta} = [\beta_1, \dots, \beta_n]$ and $\mathbf{B} = [\mathbf{b}_1, \dots, \mathbf{b}_n]^{T}$. Hence, we can denote mapping $ \mathbf{w} = \mathbf{B}\bm{\beta}$. By substituting in Eq.~\ref{eq:sw}, we get $ \bm{\beta}^T(\mathbf{B}^T\mathbf{C}_w\mathbf{B})\bm{\beta}=0$. Solution to this equation can be found by solving the following eigenvalue problem:
\begin{equation} \label{eqn:knfs}
(\mathbf{B}^T\mathbf{C}_w\mathbf{B})\bm{\beta} = 0.
\end{equation}
Solution vectors $\bm{\beta}$ can be used to calculate $\mathbf{Q}$. Since the scatter matrix $\mathbf{C}_w$ is expressed as  $\mathbf{C}_w = \frac{1}{N} \tilde{\mathbf{X}}\tilde{\mathbf{X}}^T  $,  where $\tilde{\mathbf{X}}$ is obtained by stacking zero-mean data vectors $\{\mathbf{x}_1 - \bm{\mu}, \dots, \mathbf{x}_N - \bm{\mu}\}$ as columns, Eq.~\ref{eqn:knfs} can be re-written as:
\begin{equation} \label{eqn:knfs2}
(\mathbf{B}^T\tilde{\mathbf{X}}\tilde{\mathbf{X}}^T\mathbf{B})\bm{\beta}=0.
\end{equation}
Since all terms in Eq.~\ref{eqn:knfs2} appear as product terms, a kernelized version of the problem can be obtained by replacing product terms with a kernel functions $K(\cdot, \cdot)$ following similar steps as discussed in Section~\ref{lbl:stat_features} about kernel PCA. Mapping data to a higher dimensional space through kernelization relaxes the condition that the dataset size should be small sample size is small $n \leq N$.

When formulating Null Foley--Sammon Transform in the one-class setting, the origin of the coordinate space is treated as the negative class similar to one-class SVM. Then, a single null-space vector $\mathbf{w}$ is learned considering positive data. First, mean removed training data $\tilde{\mathbf{X}}$, is projected by the null-space vector $\mathbf{w}$ to obtain vector $\mathbf{t}$. During inference, a mean removed query image $\mathbf{x}_{test} - \bm{\mu}$ is projected using  $\mathbf{w}$ to obtain $\mathbf{t}^*$.  When the difference $|\mathbf{t}-\mathbf{t}^*|$ is lower than a pre-determined threshold, the query is assigned with a positive class identity.

\subsection{Statistical Methods}
\noindent \textbf{One Class Support Vector Machine (OCSVM)}\\
One class SVM is a special case of Support Vector Machine (SVM) formulation. In a binary SVM, the  hyper-plane that separates the two classes with that largest possible margin is found. The hyper-plane is defined by support vectors. In the case of one class classification, there exists only positively labeled data during training. In One Class SVM (OCSVM), hyperplane corresponding to negative class are set to be the origin of the coordinate system \cite{Scholkopf:2001:ESH:1119748.1119749}. Therefore, the objective of OCSVM boils down to finding a hyper-plane furthest away from the origin, where positively labeled data exists in the positive half space of the hyper-plane. When this constraint is relaxed using slack variables, the optimization objective can be written as:
\begin{equation} \label{eq:ocsvm}
\begin{array}{rrclcl}
\displaystyle	\min_{\mathbf{w}, \bm{\xi}, b} & \frac{1}{2}||\mathbf{w}||^2 +\frac{1}{\nu N} \sum_i \xi_i - b \\
\text{s.t.} & \langle \mathbf{w}, \Phi(\mathbf{x}_i ) \rangle \geq b - \xi_i, \xi_i \geq 0,
\end{array}
\end{equation}
where, the column vector $\bm{\xi} = [\xi_i, \ \xi_2, \ \dots, \ \xi_N]$ and each $\xi_i$ is the slack variable corresponding to the $i^{th}$ training sample (i.e. vectorized image), $\Phi$ is a mapping function that maps $\mathbf{x}_i$ to a kernel space where dot products are defined using a kernel function $K(\cdot, \cdot)$, $b$ is the bias term and $\nu$ is a trade-off parameter, and $N$ is number of training examples. When the optimization is solved, inference on a query sample $\mathbf{x}_{test}$ can be done using the condition $\text{sgn} (\langle \mathbf{w}, \phi(\mathbf{x}) \rangle - b)$.

Eq. \ref{eq:ocsvm} can be modified with the help of Lagrange multipliers $\alpha_i, \beta_i \geq 0$ as follows:
\begin{equation}
\begin{multlined}
\mathcal{L}(\mathbf{w}, \bm{\xi}, b, \bm{\alpha}, \bm{\beta} ) = \frac{1}{2}||\mathbf{w}||^2 +\frac{1}{\nu N} \sum_i \xi_i - b \\ - \sum_i \alpha_i  ( \langle \mathbf{w}, \Phi(\mathbf{x}_i) \rangle - b + \xi_i) - \sum_i \beta_i \xi_i,  
\end{multlined}
\end{equation}
where the column vectors $\bm{\alpha} = [\alpha_i, \ \alpha_2, \ \dots, \ \alpha_N]^{T}$ and $\bm{\beta} = [\beta_i, \ \beta_2, \ \dots, \ \beta_N]^{T}$. Setting derivatives with respect to primal variables to zero, it can be shown that $\mathbf{w} = \sum_i \langle \alpha_i, \Phi(\mathbf{x}_i) \rangle$, $\alpha_i = \frac{1}{\nu N} - \beta_i \leq \frac{1}{\nu N}$  and  $\sum_i \alpha_i = 1$. By substituting these values in Eq.~\ref{eq:ocsvm}, the dual optimization problem can be derived as:
\begin{equation} \label{eq:ocsvm_dual}
\begin{array}{rrclcl}
\displaystyle	\min_{\bm{\alpha}} & \frac{1}{2}  \sum_i \sum_j \alpha_i \alpha_j K(\mathbf{x}_i, \mathbf{x}_j) \\ \text{s.t.} & 0 \leq \alpha_i \leq \frac{1}{\nu N}, \sum_i \alpha_i = 1.
\end{array}
\end{equation}
Furthermore, it can be shown that when $ 0 \leq \alpha_i \leq \frac{1}{\nu N}$ is satisfied the bias term can also be expressed as: 
\begin{equation}
b = \langle \mathbf{w}, \Phi(\mathbf{x}_i) \rangle = \sum_j \alpha_j K(\mathbf{x}_i, \mathbf{x}_j).
\end{equation}

With the dual form of the problem, as shown in Eq.~\ref{eq:ocsvm_dual}, the optimal values of parameters in problem shown in Eq.~\ref{eq:ocsvm} can be found using the kernel function $K(\cdot, \cdot)$ without explicitly defining the mapping function $\Phi(\cdot)$. The decision for any test image $x_{test}$ that is vectorized as $\mathbf{x}_{test}$ can also be expressed in terms of the kernel function using the dual variables and vectorized training images as follows:
\begin{equation}
	\text{sgn}(\sum_i \alpha_i K(\mathbf{x}_i, \mathbf{x}_{test}) - b),
\end{equation}

\noindent \textbf{Support Vector Data Descriptor (SVDD)} \\
The SVDD \cite{Tax:2004:SVD:960091.960109} formulation closely follows the OCSVM objective. However, instead of learning a hyper-plane to separate data from origin, Tax \emph{et al.} \cite{Tax:2004:SVD:960091.960109} propose to find the smallest hyper-sphere that can fit given training samples. The hyper-sphere is characterized by its mean vector (or centroid of hyper-sphere) $\mathbf{o}$ and radii $r_d>0$. The volume of hyper-sphere is minimized by minimizing $r_d \in \mathbb{R}$ while making sure hyper-sphere encloses most of the training samples. This objective can be written down in the form of optimization problem as:	
\begin{equation} \label{eq:svdd}
\begin{array}{ccclcl}
\displaystyle	\min_{\mathbf{o}, \bm{\xi}, r_d} & r_d^2 + \lambda \sum_i \xi_i \\
\text{s.t.} &  \|\mathbf{x}_i - \mathbf{o}\|^2 \leq r_d^2 + \xi_i, \xi_i \geq 0 ~ \forall i. \\
\end{array}
\end{equation}
Parameter $\lambda$ controls the trade-off between errors and the objective. Similar to the OCSVM, the above objective can be modified with the help of the Lagrangian multipliers and the updated optimization problem can be re-formulated as:
\begin{equation} \label{eq:svdd_lag}
\begin{multlined}
\mathcal{L}(r_d, \mathbf{o}, \bm{\alpha}, \bm{\gamma}, \mathbf{\xi}) = r_d^2 + \lambda\sum_i\xi_i - \sum_i\alpha_i(r_d^2 + \xi_i - \\ (\|\mathbf{x}_i\|^2 - 2 \langle \mathbf{o}, \mathbf{x}_i \rangle + \|\mathbf{o}\|^2)) - \sum_i\gamma_i\xi_i,
\end{multlined}
\end{equation}
where, the column vectors $\bm{\alpha} = [\alpha_i, \ \alpha_2, \ \dots, \ \alpha_N]^{T}$ and $\bm{\gamma} = [\gamma_i, \ \gamma_2, \ \dots, \ \gamma_N]^{T}$. By setting derivatives of primal variables to zero, it can be shown that $\sum_i \alpha_i = 1$, $\mathbf{o} = \sum_i \alpha_i \mathbf{x}_i$ and $\lambda-\alpha_i-\gamma_i = 0$. By substituting to Equation~\ref{eq:svdd_lag}, the dual form can be obtained as:
\begin{equation} 
\begin{array}{rrclcl}
\displaystyle	\min_{\bm{\alpha}} & \sum_i \sum_j \alpha_i\alpha_j \langle \mathbf{x}_i, \mathbf{x}_j \rangle - \sum_i \alpha_i\langle \mathbf{x}_i, \mathbf{x}_i \rangle\\
& \text{s.t.}  \ \ 0 \leq \alpha_i \leq \lambda, \sum_i \alpha_i = 1. \\
\end{array}
\end{equation}

A given test sample $\mathbf{x}_{test}$, is assigned a positive label if it is inside the identified hyper-sphere. More precisely, if the following condition is met:
\begin{equation} \label{eq:svdd_test}
\begin{multlined}
\|\mathbf{x}_{test} - \mathbf{o}\|^2 = \langle \mathbf{x}_{test}, \mathbf{x} \rangle - 2\sum_i\alpha_i\langle \mathbf{x}, \mathbf{x}_i \rangle \\ + \sum_i\sum_j \alpha_i\alpha_j \langle \mathbf{x}_i, \mathbf{x}_j \rangle \leq r_d^2.
\end{multlined}
\end{equation}

Since, the dual form and the inference equation both include inner product terms of $\mathbf{x}_i$ and $\mathbf{x}$, SVDD can be extended to a kernel formulation by simply replacing product terms by a kernel function that corresponds to some mapping function $\Phi$ as, $\langle \Phi(\mathbf{x}_j), \Phi(\mathbf{x}_i) \rangle = K(\mathbf{x}_i, \mathbf{x}_j)$. The kernalized version of the optimization problem of dual form then can be expressed as:
\begin{equation} 
\begin{array}{rrclcl}
\displaystyle \min_{\mathbf{\alpha} } &  \sum_i\sum_j \alpha_i\alpha_j(K(\mathbf{x}_i, \mathbf{x}_j) - \sum_i\alpha_i(K(\mathbf{x}_i, \mathbf{x}_i)))\\
& \text{s.t.} \ \ \  0 \leq \alpha_i \leq C,  \sum_i \alpha_i = 1.\\
\end{array}
\end{equation}

As we mentioned earlier that $\sum_i \alpha_i = 1$. Due to that in the case where the kernel function only depends on the difference between the two vectors, i.e., when $K(\mathbf{x}_1,\mathbf{x}_2)$ depends only on $\mathbf{x}_1-\mathbf{x}_2$, the linear term of the dual objective function becomes constant and the optimization becomes equivalent to the dual form of OCSVM in Equation~\ref{eq:ocsvm_dual} discussed in the previous section.\\

\noindent \textbf{One Class Mini-max Probability Machine (OCMPM)}\\
Similar to in one-class SVM, One class Mini-max Probability Machine (1-MPM) tries to maximize the distance between the origin and learned hyper-place with the objective of arriving at a tighter lower bound to the data. However, 1-MPM takes advantage of second order statistics of training data when the hyper-plane is learned. 
%For a given dataset with $N$ number of training images, $\mathcal{D}=\{x_i\}_{i=1}^{N}$, where each image $x_i \in \mathbb{R}^{3 \times H \times W}$. We can create a set of vectorized (as columns) images, i.e., $\{ \mathbf{x}_1, \mathbf{x}_2, \dots, \mathbf{x}_N\}$, where $\mathbf{x}_i  \in \mathbb{R}^n$ is $i^{th}$ vectorized image (as column) and $n = 3 \cdot H \cdot W$. 
In the 1-MPM algorithm, both mean $\bm{\mu}$ and the covariance matrix $\mathbf{C}$ of the vectorized images is calculated. The covariance matrix can be calculated as $\mathbf{C} = \frac{1}{N} \sum_{i=1}^{N} (\mathbf{x}_i-\bm{\mu}) (\mathbf{x}_i- \bm{\mu})^T$, where $\bm{\mu}$ is the mean is calculated as $\bm{\mu} = \frac{1}{N} \sum_{i=1}^{N}\mathbf{x}_N$. Furthermore, 1-MPM does not assume any distribution of the underlying data $\mathbf{x}$ unlike PCA (which has inherent assumption of data belonging to Gaussian distribution). With the help of both mean and covariance information of the data, 1-MPM seeks to find a hyper-plane $(\mathbf{w}, b)$ with $\mathbf{w} \in \mathbb{R}^n \setminus \{0\}, b \in \mathbb{R}$ such that data lies in the positive half space defined by $\{\mathbf{x} |\mathbf{x} \in \mathbb{R}^n, \mathbf{w}^T\mathbf{x} \geq b\}$, at least with a probability of $\tau$, even in the worst case scenario. With this background, the objective function of single class MPM can be formulated as in:
\begin{equation}
\max_{\mathbf{w} \neq 0, b} \frac{b}{\sqrt{\mathbf{w}^T\mathbf{C}\mathbf{w}}} \ \ \ \text{s.t.} ~ \inf_{(\mathbf{x}, \bm{\mu}, \mathbf{C})} P(\mathbf{w}^T\mathbf{x} \geq b) \geq \tau,
\end{equation}

when the distance between the origin and the hyper-plane is measured in terms of Mahalonabis distance with respect to
$\mathbf{C}$. Since this problem is positively homogeneous in $(\mathbf{w}, b)$ and because $\mathbf{w}\neq\mathbf{0}$ is always satisfied when $b>0$, value of $b$ is taken to be one without loss of generality. Then, a equivalent optimization problem can be obtained by minimizing the reciprocal of Mahalonobis distance as in:	
\begin{equation}\label{mpm1}
\min_{\mathbf{w}} {\sqrt{\mathbf{w}^T\mathbf{C}\mathbf{w}}}  \ \ \ \text{s.t.} ~ \inf_{(\mathbf{x}, \bm{\mu}, \mathbf{C})} \ P(\mathbf{w}^T\mathbf{x} \geq 1) \geq \tau,
\end{equation}

Using the core MPM theorem in \cite{mpm}, where it is stated that  $\inf_{(\mathbf{x}, \bm{\mu}, \mathbf{C})} \ P(\mathbf{w}^T\mathbf{x} \geq b) \geq \tau$ is equivalent to $b-\mathbf{w}^T\hat{\mathbf{x}} \geq g(\tau) \sqrt{\mathbf{w}^T \mathbf{C}}$, where $ g(\tau) = \sqrt{\frac{\tau}{1-\tau}}$, Equation~\ref{mpm1} can be re-written as:
\begin{equation} \label{mpm2}	
\min_{\mathbf{w}}   {\|\mathbf{w}^T\mathbf{C}^{\frac{1}{2}}\|_2} \ \ \ \text{s.t.} \ \ 1-\mathbf{w}^T\bm{\mu} \geq g(\tau) {\|\mathbf{w}^T \mathbf{C}^{\frac{1}{2}}\|_2}.
\end{equation}
Here it should be noted that for a real symmetric covariance matrix $\mathbf{C}$, the matrix $\mathbf{C}^{\frac{1}{2}}$ always exists. Since optimization problem shown in Equation~\ref{mpm2} is a second order cone program it can be efficiently solved using convex optimization tools.

In the robust 1-MPM formulation, it is assumed that difference between sample covariance and true covariance of the distribution will not exceed some arbitrary constant denoted as $\rho$ and Mahalanobis distance between sample mean and true mean with respect to true covariance will not exceed an arbitrary constant denoted as $\nu^2$ \cite{mpm}. Under these assumptions, it is shown in \cite{lank} that $\mathbf{w}^T\mathbf{C}\mathbf{w}$ term in Equation~\ref{mpm1} gets substituted by $\mathbf{w}^T(\mathbf{C}+\rho\mathbf{I}_n)\mathbf{w}$, while $g(\tau)$ term is changed into $K(\tau)+\nu$. \\

\noindent \textbf{Dual slope Mini-max Probability Machine (DS--OCMPM)}

Dual slope Mini-max Probability Machine is a an extension of 1-MPM considering two decision hyper-planes and availability of sub-clusters\cite{DualMPM}. In \cite{DualMPM}, a second hyper-plane $(\tilde{\mathbf{w}}, \tilde{b})$ is learned such that the data projected on the hyper-plane $\tilde{\mathbf{w}}$ has the largest possible variance. The optimization objectives takes the form of:
\begin{equation} \label{mpm3} 
\max_{\tilde{\mathbf{w}}} {\sqrt{\tilde{\mathbf{w}}^T\mathbf{C}\tilde{\mathbf{w}}}} \ \ \ \text{s.t.} ~ \inf_{\mathbf{x}, {\bm{\mu}},\mathbf{C})} \ P(\tilde{\mathbf{w}}^T\mathbf{x} \geq 1) \geq \tilde{\tau}.
\end{equation}
Since maximizing $\sqrt{\tilde{\mathbf{w}}^T\mathbf{C}\tilde{\mathbf{w}}}$ is equivalent to minimizing $\sqrt{\tilde{\mathbf{w}}^T{\mathbf{C}}^{-1}\tilde{\mathbf{w}}}$. Optimization problem in Equation~\ref{mpm3} can be transformed into another second order cone program of the form:
\begin{equation}\label{mpm4}
\min_{\tilde{\mathbf{w}}}   {\|\tilde{\mathbf{w}}^T\mathbf{C}^{-\frac{1}{2}}\|_2} \ \ \ \text{s.t.} ~ 1-\tilde{\mathbf{w}}^T\bm{\mu} \geq g(\tilde{\tau}) {\|\tilde{\mathbf{w}}^T\mathbf{C}^{-\frac{1}{2}}\|_2}.
\end{equation}
Assuming that the difference between sample covariance and true covariance of the distribution will not exceed some arbitrary constant $\tilde{\rho}$ for $\mathbf{C}^{-1}$, the robust version of the optimization problem is obtained by substituting $\tilde{\mathbf{w}}^T{\mathbf{C}}^{-1}\tilde{\mathbf{w}}$ term in Equation~\ref{mpm3} by $\tilde{\mathbf{w}}^T{(\mathbf{C}+\tilde{\rho}\mathbf{I}_n)}^{-1}\tilde{\mathbf{w}}$.

In order to mitigate the effect of sub-distributions, data is clustered into $k$ clusters using the Ward's method \cite{ward1963hierarchical}. Global mean and variance $(\bm{\mu}, \mathbf{C})$ are calculated with respect to the whole dataset along with cluster-wise statistics $(\bm{\mu}_i, \mathbf{C}_{i})$ for $i^{th}$ cluster. Optimization is carried out over cumulative covariance of each individual cluster $\sum_{i} \sqrt{\mathbf{w}^T\mathbf{C}_{i}\mathbf{w}}$ while constraints are defined with respect to global statistics $(\bm{\mu}, \mathbf{C})$. 

Given global and local clusters $(\bm{\mu}, \mathbf{C})$ and $(\bm{\mu}_i, \mathbf{C}_{i})$ for $i=1, 2, \dots, k$, where $k$ is the number of clusters, the following joint-optimization problem is solved to find both $\mathbf{w}$ and $\tilde{\mathbf{w}}$:
\begin{equation}
\begin{aligned}
& \underset{\mathbf{w}, \tilde{\mathbf{w}}}{\text{minimize}}
& & \sum_{i=1}^{k} ||\mathbf{w}^T(\mathbf{C}_{i} + \rho\mathbf{I}_n)^{\frac{1}{2}}||_2 + \|\tilde{\mathbf{w}}^T(\mathbf{C}_{i} + \tilde{\rho}\mathbf{I}_n)^{-\frac{1}{2}}\|_2 \\
& \text{subject to}
& & (g(\tilde{\tau})+\nu){||\tilde{\mathbf{w}}^T(\mathbf{C}_{i} + \tilde{\rho}\mathbf{I}_n)^{-\frac{1}{2}}||_2} - 1 \leq \tilde{\mathbf{w}}^T\bm{\mu}\\
&&& (g(\tau)+\nu){\|\mathbf{w}^T(\mathbf{C}_{i} + \rho\mathbf{I}_n)^{\frac{1}{2}}\|_2} - 1 \leq \mathbf{w}^T\bm{\mu}.
\end{aligned}
\end{equation}
Since the product of $\mathbf{w}$ and $\tilde{\mathbf{w}}$ do not appear in the optimization statement, this problem can be solved independently for $\mathbf{w}$ and $\tilde{\mathbf{w}}$ using two second order cone programs. Once hyper-plane parameters are obtained, given a test sample $\mathbf{x}_{test}$, identity of the sample would be assigned to be negative if $\mathbf{w}^T\mathbf{x}_{test} < 1 \cap \tilde{\mathbf{w}}^T\mathbf{x}_{test} < 1$, and positive otherwise.\\

\begin{figure*}
	\centering
	\includegraphics[width=1\linewidth]{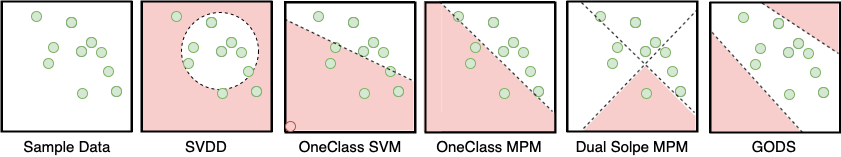}
	\caption{ Decision boundaries obtained by different statistical one class classifiers.}
	\label{fig:comp}
\end{figure*}

\noindent \textbf {Generalized One-class Discriminative Sub-spaces (GODS)}\\
GODS \cite{Wang_2019_ICCV} extends one-class SVM formulation into two separating hyper-planes. Similar to one-class SVM, the first hyper-plane $(\mathbf{w}, b)$ is learned such that most of data points appear in the positive half space defined by the hyper-plane. In addition, the second hyper-plane $(\tilde{\mathbf{w}}, \tilde{b})$ is learned such that most of the data lie in the negative space defined by the hyper-plane. A basic variant of the GODS classifier can be learned using the following optimization objective:
\begin{equation} 
\begin{array}{rrclcl}
\displaystyle	\min_{(\mathbf{w}, b), (\tilde{\mathbf{w}}, \tilde{b}), \bm{\xi}, \tilde{\bm{\xi}}, \beta > 0} & \frac{1}{2}||\mathbf{w}||^2 + \|\tilde{\mathbf{w}}\|^2 \\  & - b - \tilde{b} + \lambda \sum_i \xi_{i} + \tilde{\mathbf{\xi}_i} \\
\text{s.t.} & \mathbf{w}^T\mathbf{x}_i - b \geq \eta - \xi_{i}, \\
& \tilde{\mathbf{w}}^T\mathbf{x}_i - \tilde{b} \leq \eta - \tilde{\xi}_{i}, \\
& dist((\mathbf{w}, b), (\tilde{\mathbf{w}}, \tilde{b})) \leq \beta,
\end{array}
\end{equation}
where,  $\lambda$ is the slack regularization constant, $\eta$ defines the classifier margin and $dist(\cdot, \cdot)$ is suitable distance function between the two hyper-planes.  This loss function forces to find the most similar pair of hyper-planes that satisfies the stated objective. The additional constraint on distances prevents from finding a trivial solution. By explicitly setting $\|\mathbf{w}\|_2^2 = 1$ and $\|\tilde{\mathbf{w}}\|_2^2 = 1$ and setting $dist(\cdot, \cdot)$ to be the Euclidean distance, this objective can be simplified as:
\begin{equation} \label{eq:gods1}
\begin{array}{rrclcl}
\displaystyle & \underset{b, \tilde{b}, \bm{\xi}, \bm{\xi}, \beta > 0}{\min} \ \ \ \ \  g(b, \tilde{b}) - 2 \ \mathbf{w}^T\tilde{\mathbf{w}} + \lambda \sum_i \xi_{i} + \tilde{\xi}_{i} \\
& \text{s.t.} \ \ \ \ \ \ \ \ \ \ \ \ \ \sum_{i} \lfloor \eta - (\mathbf{w}^T(\mathbf{x}_i + b) - \xi_{i} \rfloor_+ \\
& +  \lfloor \eta - (\tilde{\mathbf{w}}^T(\mathbf{x}_i + \tilde{b}) - \tilde{\xi}_{i} \rfloor_+, \\
& \|\mathbf{w}\|_2^2 = 1, \|\tilde{\mathbf{w}}\|_2^2 = 1, \\
\end{array}
\end{equation}
where $g(b, \tilde{b}) = (b-\tilde{b})^2-b-\tilde{b}$ and notation $\lfloor \rfloor_+$ denotes hinge loss. This initial formulation can be extended to a generalized formulation by introducing sub-spaces in place of hyper-planes. Let $\mathbf{W}, \tilde{\mathbf{W}} \in S^M_n$ be orthonormal  subspace frames, where $\mathbf{W}^T\mathbf{W} = \tilde{\mathbf{W}}^T\tilde{\mathbf{W}} = \mathbf{I}_M$. Here, $\mathbb{I}_M$ is identity matrix of size $M \times M$. Such frames  $S^M_n$ with $n$ dimensional subspaces, belong to the Stiefel manifold and each column in the matrices $\mathbf{W}, \tilde{\mathbf{W}}$ of size $n \times M$ is orthonormal to the rest. The number of hyper-planes $M$ is a hyper-parameter and can be set differently for each dataset to achieve better performance. By replacing the distance term $dist((\mathbf{w}, b), (\tilde{\mathbf{w}}, \tilde{b}))$ by the Euclidean distance of each data point from both hyper-planes, Equation~\ref{eq:gods2} can be modified as:
\begin{equation} \label{eq:gods2}
\begin{array}{llllll}
\displaystyle & \underset{\mathbf{W}, \tilde{\mathbf{W}}, \mathbf{b}, \tilde{\mathbf{b}}, \bm{\xi} > 0}{\min} \ \frac{1}{2} \sum_{i=1}^N (\|\mathbf{W}^T\mathbf{x}_i+\mathbf{b}\|_2^2 + \|\tilde{\mathbf{W}}^T\mathbf{x}_i+\tilde{\mathbf{b}}\|_2^2) \\
& \ \ \ \ \ \ \ \ \ \ \ \ \ \ \ \ + \ g(\mathbf{b}, \tilde{\mathbf{b}}) + \lambda \sum_i (\xi_{i} + \tilde{\xi}_{i}) \\
& \ \ \ \ \ \ \ \ \ \ \ \ \ \ \ \ + \ \frac{\nu}{N} \sum_i \lfloor\eta - \min(\mathbf{W}^T(\mathbf{x}_i + \mathbf{b}) - \xi_{i}\rfloor^2_+\\
& \ \ \ \ \ \ \ \ \ \ \ \ \ \ \ \ + \ \frac{1}{2N}  \sum_i \lfloor\eta + \max(\tilde{\mathbf{W}}^T(\mathbf{x}_i + \tilde{\mathbf{b}}) - \tilde{\xi}_{i} \rfloor^2_+.
\end{array}
\end{equation}
Classifier obtained by solving this optimization is named as the \textit{generalized one-class discriminative subspace} (GODS) classifier. \cite{Wang_2019_ICCV} presents a parameter initialization method and an  efficient optimization method to optimized the proposed optimization objective.

Figure~\ref{fig:comp} illustrates a pictorial comparison of the decision spaces obtained by different statistical one-class classifiers.\\

\subsection{Deep Learning Methods}

\subsubsection{Discriminative Methods}
The discriminative methods utilize loss functions that are typically inspired from the statistical methods like OC-SVM and SVDD or utilize regularization techniques to make traditional neural network training more compatible to one-class classification.\\

\begin{figure}
	\centering
	\includegraphics[width=0.7\linewidth]{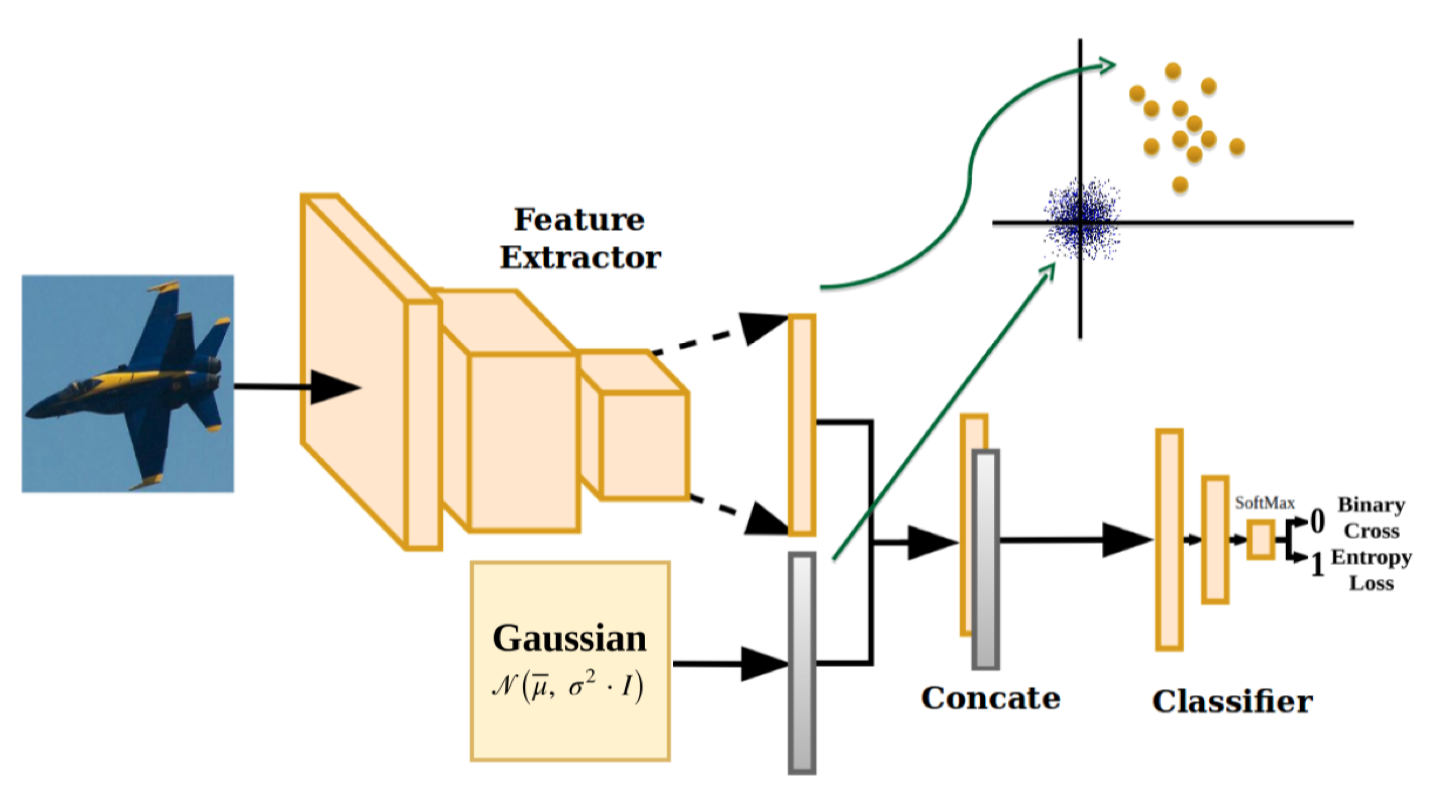}
	\caption{ Training strategy used in One-Class CNN \cite{oza2019one} is inspired from OC-SVM training. Here, the features are trained to be different compared to zero-centered Gaussian distribution, same as OC-SVM learns separating hyper-plane between one-class data from the origin.}
	\label{fig:ocnn}
\end{figure}

\noindent \textbf {One Class CNN (OCCNN)}\\	
One class CNN is a deep learning based classifier inspired by One Class SVM \cite{oza2019one}. It comprises of two sub-networks -- a feature extractor and a classifier as shown in Figure~\ref{fig:ocnn}. The feature extractor network is learned using an OOD dataset in \cite{oza2019one}. However, in theory any One Class feature learning method  can be used for this purpose. The classifier is a fully connected network that terminates with two outputs corresponding to positive and negative classes.

During training, positively labeled data are passed through the network to first obtain features. These  features are assigned a class label $y=0$. Then, a batch of random noise is sampled from the feature space using a Gaussian distribution $\mathcal{N}(0,\sigma^2\textbf{I})$. This batch is assigned a class label of $y=1$. Two batches are concatenated and passed through the classifier to obtain a prediction. The network is trained using binary cross entropy loss defined as:
\begin{equation}
\mathcal{L}  = - \frac{1}{2k} \sum_{j=1}^{2k} y \log(p_0) + (1-y) \log(1-p_1),
\end{equation}
where $k$ is the batch size, and $p_0, p_1$ are the outputs corresponding to the zeroth and first classes in the network respectively ($1-p_0 = p_1$). This work was later extended in \cite{oza2019active} and \cite{Yash_IJCB2020} for active authentication and face presentation attack detection, respectively.\\	

\noindent \textbf {Deep SVDD (DSVDD)}\\	
The deep SVDD framework introduced in \cite{dsvdd} learns a deep representation with the objective of enclosing embeddings of the positively labeled data with the smallest possible hyper-sphere.  Let $F$ be the network function of the deep network parameterized by weights $\theta_f$ consisting of $L$ layers and $l^{th}$ layer parameters denoted as $\theta_f^l$. When the center and the radius of the hyper-sphere is defined by $\mathbf{o}$ and $r_d$ respectively, the objective of the network is to learn a representation that minimize the loss given as:
\begin{equation}
\begin{multlined}
\min_{r_d, \theta_f} \ \ r_d^2 \ + \ \frac{1}{\nu N} \sum_{i=1}^N \max\{0, \|F({x}_i) - \mathbf{o}\|^2 - r_d^2\} \\ 
+ \frac{\lambda}{2} \sum_{l=1}^{L} \|\theta_f^l\|_F^2,
\end{multlined}
\end{equation}
The loss function comprises of three loss terms. The first term encourages the network to find a hyper-plane with a smaller radii. The second term penalizes points lying outside the hyper-sphere. The third term is weight decay added to network parameters. Parameter $\nu$ is shown to be an upper bound on the fraction of outliers and a lower bound on the fraction of samples being outside or on the boundary of the hypersphere.

A second loss function is proposed for the case when most of the training data belongs to the positive class (which is the typical setting for One Class Classification). The second loss function is defined as:
\begin{equation}
\begin{multlined}
\min_{\theta_f} \frac{1}{N} \sum_{i=1}^N  \|F({x}_i)-\mathbf{o}\|^2 + \frac{\lambda}{2} \sum_{l=1}^{L} \|\theta_f^l\|_F^2,
\end{multlined}
\end{equation}

In this loss, the authors are merely forcing the embedding of the positive class to have a lower intra-class variance similar to \cite{ocfeatures}. During inference, for a given query sample $x_{test}$, a positive class identity is declared if $\|F({x}_{test})-\mathbf{o}\|^2$ is smaller than a selected threshold. Furthermore, Ruff \emph{et al.} \cite{dsvdd} found when $\mathbf{o}$ is set to be a learnable parameter, network leads to a trivial solution. Therefore, they set $\mathbf{o}$ to be the mean of outputs obtained at the  initial forward pass.  They further found that, using bias terms in the network would also lead to a trivial solutions. Therefore, they do not use any bias terms in the network architecture. Finally, authors advise against bounded activation functions in the network as it too could contribute to trivial solutions.\\

\noindent \textbf{Holistic Approach to One Class (HRN)}\\	
HRN proposes \cite{hu2020hrn} to train a deep classification network trained with log-likelihood loss with regularization applied on the learned features. The proposed loss is termed as holistic regularization or  H-regularization. Furthermore, HRN adds a 2-norm instance level data normalization strategy to deal with different feature scales in the data instances. Given one-class data sampled from distribution $p_{{x}}$ the training loss for HRN can be expressed as:
\begin{equation}
\mathcal{L}=\underbrace{\underset{{x} \sim p_{{x}}}{\mathbb{E}}[-\log (\operatorname{Sigmoid}(\mbox{En}({x})))]}_{\text {NLL }}+\lambda \underbrace{\underset{{x} \sim p_{{x}}}{\mathbb{E}} [ \ \left\|\nabla_{{x}} \mbox{En}({x})\right\|_{2}^{h}}_{\text {H-regularization }} \ ],
\end{equation}

Minimizing the negative log-likelihood (NLL) helps encoder network $\mbox{En}$ to model the one-class training data distribution. Here, the holistic regularization (or H-regularization) is added to control the output of the encoder network that might saturate the sigmoid function value and remove the feature bias introduced due to input data with high values that might lead to unimportant features. On top of having these two loss functions, HRN also adds a 2-norm instance-level data normalization that aims to solve problems caused due to different feature scales which might confuse model to have poor performance \cite{hu2020hrn}. In 2-norm instance normalization, the data ${x}$ normalized such that it has $\|{x}\|_2=1$. Furthermore, mean value from each feature is subtracted to make sure feature values of each instance have zero-mean.  This proposed normalization strategy encourages some parameters in the encoder network to be negative and consequently it increases the value space for probability of learning a better encoder model. The proposed normalization is shown to work really well for one-class classification compared to commonly used instance normalization \cite{ulyanov2016instance}.

\subsubsection{Generative Models}
The generative models are one of the most popular methods for one-class learning. These models are based on a range of methods such as de-noising auto-encoders, generative adversarial networks, auto-regressive models, adversarial auto-encoders etc.\\

\noindent \textbf {AnoGAN }\\	
AnoGAN proposes to learn a Generative Adversarial Network (GAN) based on positively labeled data \cite{IPMI}. Generative Adversarial Network is a type of a generative model that consists of two sub-networks - a Generator ($G$) and a Discriminator ($D$). The set of images used to train a GAN is referred  as \textit{real images}. The goal of the generator network is to use a random noise vector $\mathbf{z}$ to generate images that closely resemble \textit{real images}. Images generated by the generator is referred to as \textit{fake images}. The goal of the discriminator is to differentiate between \textit{real images} from  \textit{fake images}. 

In order to achieve this objective, discriminator is designed to generate a high score for \textit{real images} and a low score for \textit{fake images}. Therefore, discriminator parameters are learned such that  $\log D(x)$ and $\log (1-D(G(\mathbf{z})))$ are maximized, where $x$ and $\mathbf{z}$  are \textit{real image} samples and random noise vectors, respectively. Therefore, optimization in discriminator update becomes:
\begin{equation}
\min_G \max_D \mathbb{E}_{{x} \sim p_{x}} [\log D({x})]+\mathbb{E}_{\mathbf{z} \sim p_\mathbf{z}} [\log (1-D(G(\mathbf{z})))],
\end{equation}

Given a query ${x}_{test}$, it is required to obtain the corresponding latent mapping $\mathbf{z}$ for inference. In order to approximate the corresponding latent mapping, first starting from a random vector $\mathbf{z}$, its value is updated by  back-propagating the loss as:
\begin{equation}
\begin{array}{llllll}
\mathcal{L}(\mathbf{z}) & = (1-\lambda) \sum |{x}_{test} - G(\mathbf{z})|  \\
& \ \ \ \ \ \ \ \ + \lambda  \sum |Fe_D({x}_{test}) - Fe_D(G(\mathbf{z}))|,
\end{array}
\end{equation}
where $Fe_D$ is a feature extracted from the Discriminator. Here, $\mathbf{z}$ is the approximated latent vector and $G(\mathbf{z})$ is the corresponding fake image. The first term of the loss minimizes the difference between real and fake image in the pixel space where as the second term minimizes the difference in feature space. The total loss reduces when the approximated latent vector approaches the corresponding  latent vector of input ${x}$. After $T$ updates, the loss $\mathcal{L}$ is treated as the novelty score of the input image. If it is lower that a selected threshold it is assigned a positive label.\\

\noindent \textbf {Adversarially Learned One-Class Classifier (ALOCC) }\\	
The main challenge in AnoGAN is to find the corresponding latent vector given a query input. To circumvent this challenge, a conditional GAN was used in \cite{cvpr2018}. A conditional GAN, similar to a GAN, has a Generator and a Discriminator. However, different from a GAN, a conditional GAN accepts an image  and  a noise vector as the input. The Generator learns  to generate images (fake images) that have a close resemblance to real images. Discriminator tries to differentiate real images from fake images same as in a GAN.

\begin{figure}
	\centering
	\includegraphics[width=0.85\linewidth]{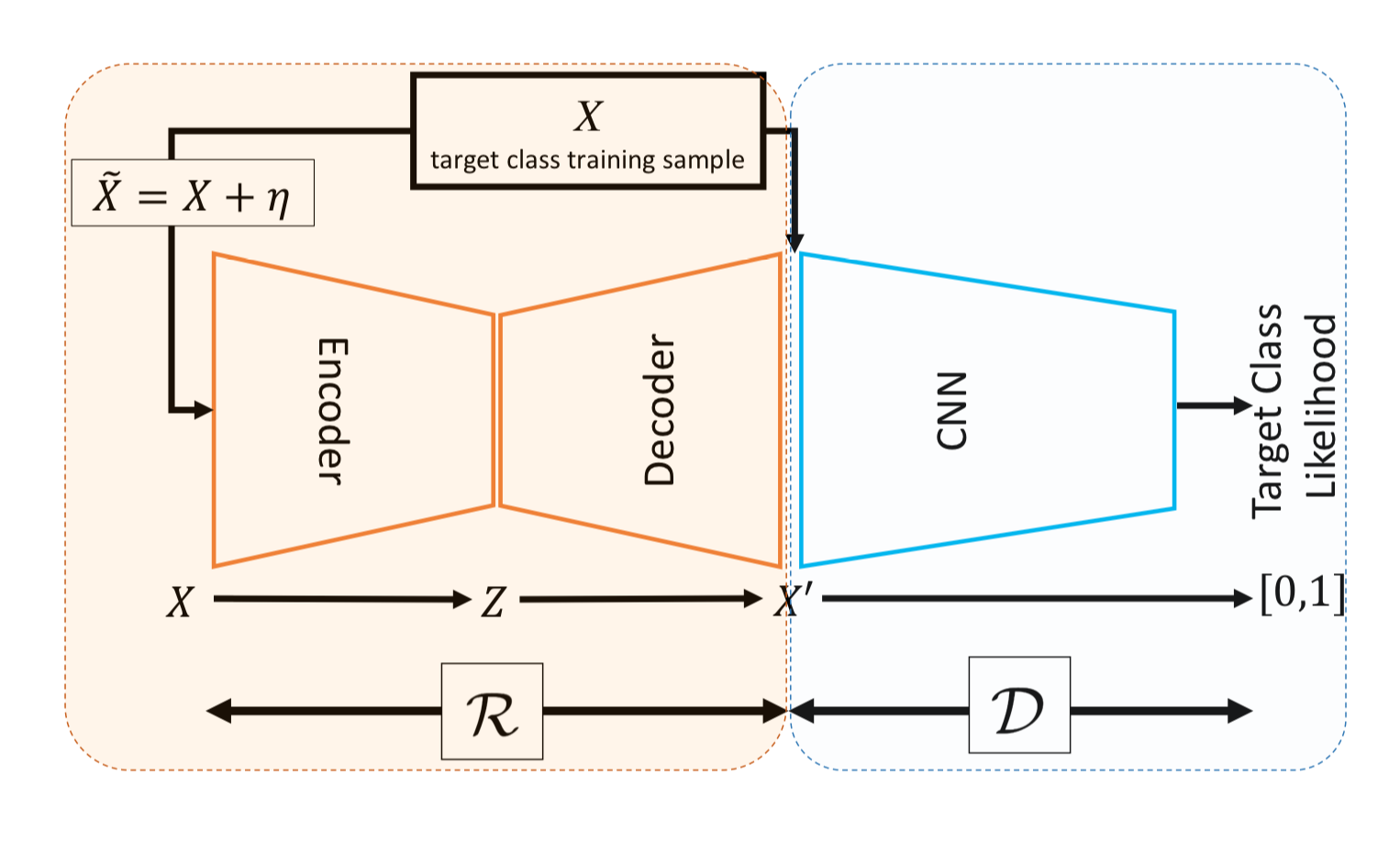}
	\vskip -2.5pt
	\caption{ALOCC utilizes a de-noising auto-encoder network architecture \cite{cvpr2018} trained with generative adversarial training. Here, the input images are corrupted with Gaussian noise and network is expected to reconstruct input image perfectly.}
	\label{fig:alooc}
\end{figure}

The network architecture proposed in  \cite{cvpr2018} is shown in Figure~\ref{fig:alooc}. A auto-encoder is used as the Generator (denoted by $\mathcal{R}$), and a CNN is used as the Discriminator (denoted by $\mathcal{D}$). During training, Gaussian noise ${\xi}$ is added to the input ${X}$ and the noisy input image ${\hat{x}} = {X}+{\xi}$  is given as the input to the Generator. The Generator tries to produce denoised outputs that corresponds to $\mathbf{X}$ . The Discriminator tries to differentiate noise-less images $\mathbf{X}$ from denoised outputs $\mathcal{R}({X}+{\xi}  )$. The network is trained using GAN loss defined as:
\begin{equation}
\min_\mathcal{R} \max_\mathcal{D} \mathbb{E}_{{x} \sim p_{x}} [\log D({x})]+\mathbb{E}_{{\hat{{x}}}\sim p_{{x}+{\xi}} }[\log (1-D(G({\hat{x}})))],
\end{equation}
During inference, given a query image $\mathbf{x}$, it is declared a positive query if $\mathcal{D}( \mathcal{R} ({x})  )$ is greater than a pre-determined threshold.\\

\noindent \textbf {Old is Gold (OGN)}\\
Old is Gold is an extension of the ALOCC framework \cite{zaheer}. It consists of the same architecture as ALOCC \cite{cvpr2018} withe a Generator network ($\mathcal{G}$) and a Discriminator network ($\mathcal{D}$) as shown in Figure~\ref{fig:oldisgold}.  The network is first pre-trained using the ALOCC method .  Then, \cite{zaheer} proposes to fine-tune the  network by providing two different types of \textit{fake} images -- namely \textit{bad quality examples }${x}_{low}$ and \textit{pseudo anomaly images} ${x}_{pseudo}$.

In order to generate  \textit{bad quality examples }${x}_{low}$, an older snapshot of the Generator network $\mathcal{G}_{old}$ is used.  A \textit{pseudo anomaly image} is generated considering the interpolation of two \textit{bad quality examples } in the image space as:
\begin{equation}
{x}_{pseudo} = \frac{\mathcal{G}_{old}({x}_i) + \mathcal{G}_{old}({x}_j)}   {2},
\end{equation}
where $i \neq j$. Both  \textit{bad quality examples } and \textit{pseudo anomaly image} simulate unusual inputs. Therefore, authors argue that when the Discriminator is trained to differentiate these unusual inputs from normal inputs (${x}$ and ${\hat{x}}$), the Discriminator can be used to effectively identify positive samples during inference. Therefore, during the fine-tuning stage, the Discriminator parameters are optimized such that the Discriminator is able to differentiate real image and fake images from \textit{bad quality examples } and \textit{pseudo anomaly images}. The optimization objective is defined as:
\begin{figure}[!b]
	\centering
	\includegraphics[width=0.85\linewidth]{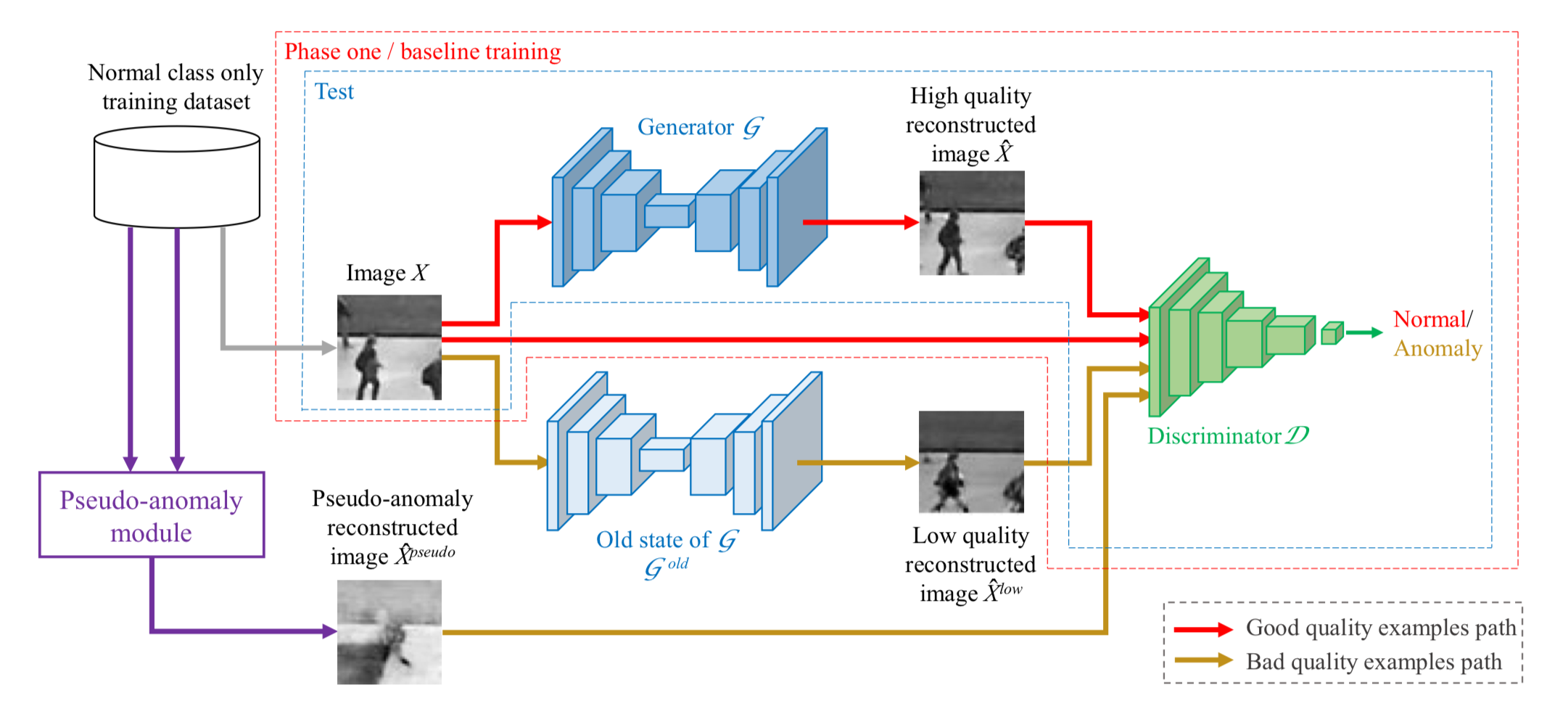}
	\vskip -2.5pt
	\caption{In old is gold method training, the old generator parameters are used to create a pseudo-negative images that acts as a negative data to build the one-class classifier that can identify underlying one-class data \cite{zaheer}.}
	\label{fig:oldisgold}
\end{figure}
\begin{equation}
	\begin{multlined}
\max_\mathcal{D}  \alpha \mathbb{E}_{{x} \sim p_{x}} [\log (1-D({x}))] \\ + (1- \alpha) \mathbb{E}_{\mathbf{\hat{{x}}}\sim p_{{x}+{\xi}} }[\log (1-D(G({\hat{x}})))] \\ +  \beta \mathbb{E}_{{x} \sim p_{{x}_{low}}} [\log (D({x}))] \\ +(1-\beta) 	\mathbb{E}_{{\hat{{x}}}\sim p_{{x}_{pseudo}} }[\log (D({x}))],
 	\end{multlined}
\end{equation}
\begin{figure*}[!t]
 	\centering
 	\includegraphics[width=1\linewidth]{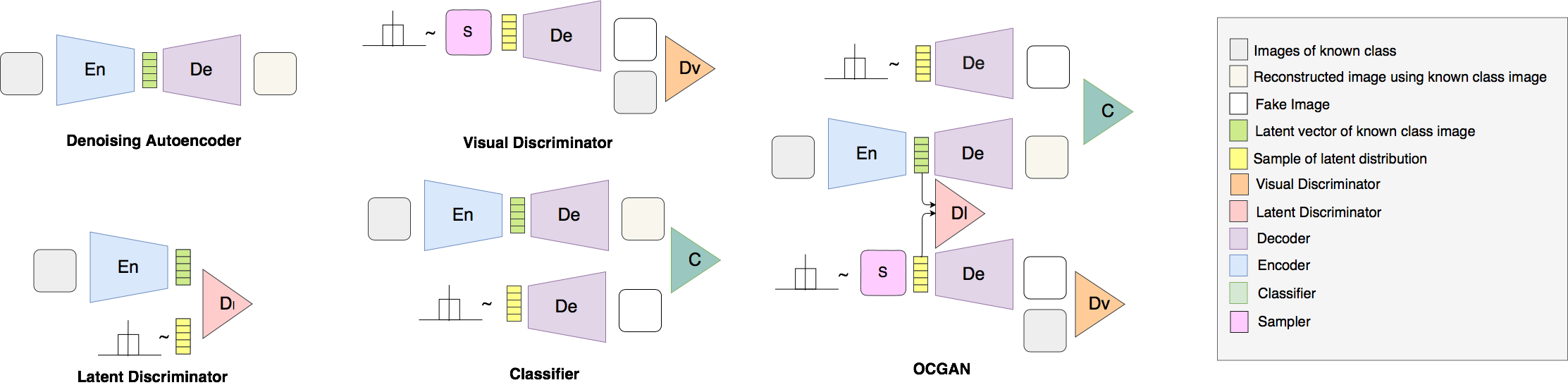}
 	\caption{The figure shows different component of OCGAN network architecture used to learn one-class classifier \cite{ocgan}. It consists of an auto-encoder network, one discriminator for image, one discriminator for latent space and one classifier to guide image discriminator update through informative sampling.}
 	\label{fig:ocgan}
\end{figure*}
where $\beta$ and $\alpha$ are hyper-parameters. During inference, a given image ${x}$ is assigned to the positive class if $\mathcal{D} ( \mathcal{G}({x}))$ is smaller than a predetermined threshold.\\

\noindent \textbf {One Class GAN (OCGAN)}\\			
OCGAN \cite{ocgan} is a representation based classifier that uses an denoising auto-encoder (with Encoder $\mbox{En}$ and decoder $\mbox{De}$) backbone to learn a representation. Authors point out that when the positive class is sufficiently diverse, auto-encoders learn a generic set of filters that leads to  good representations for all types of inputs. When this is the case, reconstruction based novelty detection fails. 

In order to mitigate this effect, authors propose to first learn a representation such that the latent distribution of input images follow a uniform distribution, i.e. $\mathcal{U}(-1,1)$. This is done adversarially using a \textit{Latent Discriminator}. Then, latent vectors are sampled from  $\mathcal{U}(-1,1)$ and its corresponding decoded output is forced to follow the distribution of real images. This is achieved adversarially using a  \textit{Visual Discriminator}. The auto-encoder 
network functions as the Generator during training. Once training converges, every latent code is guaranteed to produce an image from the positive class. Therefore, reconstruction error for non-positive images becomes high.

Three types of images are used in OCGAN training. Images from the training dataset ${x}$ are called \textit{real images}. Outputs of the denoising auto-encoder $\mbox{De}(\mbox{En}({x} + {n}))$ where ${n}$ is the added noise, are referred to as \textit{reconstructed images}. Finally, generated images from random $\mathcal{U}(-1,1)$ noise $\mbox{De}(\mathbf{z}), \mathbf{z} \sim \mathcal{U}(-1,1)  $ are called \textit{fake images}. OCGAN model contains four sub-networks as shown in Figure~\ref{fig:ocgan}.

\begin{itemize}
	\item \textbf{Denoising auto-encoder}. The encoder is terminated with a tanh activation, forcing the latent space to dimensions to have a range $(-1,1)$. It maps a noisy input to the latent space and learns a inverse mapping back to the image space. auto-encoder is learned using the loss $$ \mathcal{L}_{\mbox{MSE}} = \|{x} - \mbox{De}(\mbox{\mbox{En}}({x}+{n})) \|_2^2.$$
	\item \textbf{Latent discriminator}. Learns to discriminate between encoded real images and random samples generated from $\mathcal{U}(-1,1)$. Parameters are learned by minimizing GAN loss \begin{eqnarray}
	\mathcal{L}_{\mbox{latent}}  &&= -(\mathbb{E}_{s \sim \mathbb{U}(-1,1) } [ \log D_l(s) ] + \nonumber\\
	&& \mathbb{E}_{x \sim p_x}[\log (1-D_l( \mbox{\mbox{En}}(x+n)))]). \nonumber
	\end{eqnarray}
	\item \textbf{Visual discriminator}. Learns to discriminate between fake images and real images. \begin{eqnarray}
	\mathcal{L}_{\mbox{visual}}  &=& -(\mathbb{E}_{s \sim \mathbb{U}(-1,1) } [ \log D_v(\mbox{De}(s)) ] + \nonumber\\
	&& \mathbb{E}_{x \sim p_l}[\log (1-D_v( x))]). \nonumber
	\end{eqnarray}
	\item \textbf{Classifier}. Learns to recognize real images from fake images. Classifier is trained using the cross entropy loss. It is used to guide the sampling process such that informative samples are chosen during generator update.
\end{itemize}

\begin{figure}[!b]
	\centering
	\includegraphics[width=1\linewidth]{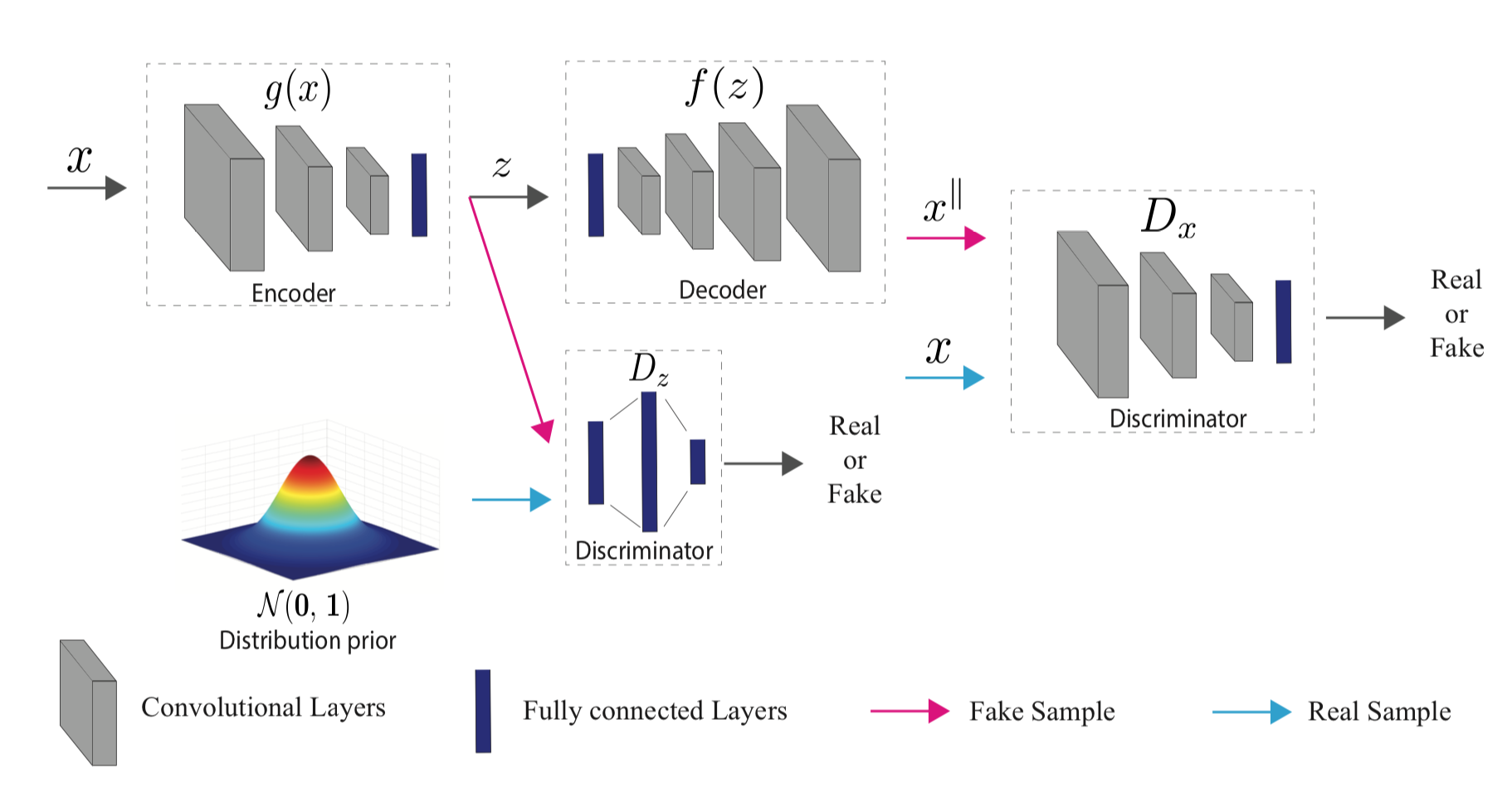}
	\vskip -2.5pt
	\caption{GPND method follows training strategy of adversarial encoder \cite{GPND}, where the latent space is forced to be standard normal distribution through latent discriminator and auto-encoder is trained to produce high-quality images through visual discriminator.}
	\label{fig:gpnd}
\end{figure}

First, for a given input ${x}$, keeping everything else fixed, the classifier loss is evaluated and classifier weights are updated. Secondly, two discriminator losses are evaluated ($\mathcal{L}_{\emph{latent}}+\mathcal{L}_{\emph{visual}} $) and weights of the two discriminators are updated accordingly.  Then, negative mining is carried out in the latent space by using classifier output for guidance. Specifically, the latent sample is changed using back-propagation such that it may fool the learned classifier. Finally, the resulting latent vectors are used to evaluate $(\mathcal{L}_{\emph{latent}}+\mathcal{L}_{\emph{visual}}+\lambda \mathcal{L}_{\emph{mse}})$ which is used to update Generator weights.\\

\noindent \textbf {Generative Probabilistic Novelty Detection (GPND)}\\
GPND\cite{GPND} proposed a probabilistic framework based solution for one class classification by modeling the data manifold of positive data. Authors first define a functional mapping between the latent space and data and then linearizes the mapping function around the data point. It is shown that the probability is factorized with respect to local coordinates of the manifold tangent space.

In \cite{GPND}, the  relationship between a training instance ${x}_i$ and its latent variable $\mathbf{z}_i$ is modeled as  ${x}_i = f(\mathbf{z}_i) + \mathbf{\zeta}_i$, where $\mathbf{\zeta}_i$ is additive noise. Let $g$ be the inverse process of $f$ defined as $f(g({x}_i)) = \mathbf{z}_i$.

For a new data sample $\mathbf{\bar{x}}$ with $\mathbf{\bar{z}} = g(\mathbf{\bar{x}})$, assuming $f$ is smooth, linearization can be performed based on first order Taylor expansion as follows:
\begin{equation}
	f(\mathbf{z}) =  f(\mathbf{\bar{z}}) + J_f (\mathbf{\bar{z}}) (\mathbf{z} - \mathbf{\bar{z}}) + O(\| \mathbf{z} - \mathbf{\bar{z}}\|^2),
\end{equation}
where $J_f (\mathbf{\bar{z}})$ is the Jacobi matrix computed at $ \mathbf{\bar{z}}$. The tangent space of $f$ at ${x}^\parallel$ is spanned by $\mathbf{U}^\parallel$, where $J_f(\mathbf{\bar{z}}) = \mathbf{U^\parallel S V}^T$ is the SVD decomposition. Data point ${\bar{x}}$ can be expressed with respect to local coordinates that defines the tangent space and the space orthogonal to it as follows:
\begin{equation}
\mathbf{\bar{w}} = \mathbf{U}^T {\bar{x}} = \begin{bmatrix}
\mathbf{U^\parallel}^T {\bar{x}} \\
\mathbf{U^\perp}^T {\bar{x}}
\end{bmatrix} =  \begin{bmatrix}
\mathbf{\mathbf{\bar{w}^\parallel} } \\
\mathbf{\mathbf{\bar{w}^\perp} } 
\end{bmatrix},
\end{equation} 

Assuming  $ \mathbf{\mathbf{\bar{w}^\parallel} }$ and $\mathbf{\mathbf{\bar{w}^\perp} }  $ are independent, $p_{{x}}({{x}})$ can be written as:
\begin{equation}
p_{{x}}({{x}}) =  p_{\mathbf{w}}({\mathbf{w}}) =   p_{\mathbf{w}^\parallel}(\mathbf{\mathbf{\bar{w}}^\parallel} )   p_{\mathbf{w}^\perp}(\mathbf{\mathbf{\bar{w}}^\perp} ),
\end{equation}
For a given test data point ${x}$, \cite{GPND} shows that$ \mathbf{\mathbf{\bar{w}^\parallel} }$ and $\mathbf{\mathbf{\bar{w}^\perp} }  $  can be approximated using following formula:
\begin{equation}
p_{\mathbf{w}^\parallel}(\mathbf{\mathbf{\bar{w}}^\parallel} ) = | det S^{-1}| p_{\mathbf{z}}(\mathbf{z}),
\end{equation}
\begin{equation}
p_{\mathbf{w}^\perp}(\mathbf{\mathbf{\bar{w}}^\perp} ) \approx \frac{\Gamma(\frac{m-k}{2})}{2 \pi^ {\frac{m-k}{2} }\| \mathbf{w}^\perp\|^{m-k}} p_{\|\mathbf{w}^\perp\|}(\mathbf{\|\mathbf{\bar{w}}^\perp\|} ),
\end{equation}
where $k$ is the dimension in the latent space and $\Gamma$ is the Gamma function. The distribution of norms $p_{\|\mathbf{w}^\perp\|}$ is learned offline by histogram norms of $\mathbf{w}^\perp =  {\mathbf{U}^\perp}^T  \mathbf{\bar{x}}$. During inference, this probability is calculated, and if the probability is greater than a pre-determined threshold it is declared to be a positive sample.

Functions $f$ and $g$ are learned using a adversarial auto-encoder network, where  $f$ and $g$ represents the encoder and decoder functions of the network respectively as shown in Figure~\ref{fig:gpnd}. The proposed network has two discriminators. The first Discriminator $ D_z $ operates on  the latent space that forces encoder embeddings to follow $\mathcal{N}(0,\textbf{I})$ distribution. The second Discriminator  $ D_x$  operates in the image space and forces output of the auto-encoder to follow the distribution of real data. The network is trained adversarially by optimizing over:
\begin{equation}
\hat{g}, \hat{f} = \arg \min_{g,f} \max_{D_x, D_z} L_{adv-d_z} + L_{adv-d_x} + \lambda L_{error},
\end{equation}
where,
\begin{equation}
\mathcal{L}_{adv-d_z} = \mathbb{E} [\log(D_z(\mathcal{N}(0,\textbf{I})))] + \mathbb{E}[\log(1-D_z(g({x})))],
\end{equation}
\begin{equation}
\mathcal{L}_{adv-d_x} = \mathbb{E}[\log(D_x({x}))] + \mathbb{E}[\log(1-D_x(f(\mathcal{N}(0,\textbf{I}))))],
\end{equation}
\begin{equation}
\mathcal{L}_{error} = -\mathbb{E}_z[\log(p(f(g({x}))|{x}))],
\end{equation}

\begin{figure}[htp!]
	\centering
	\includegraphics[width=1\linewidth]{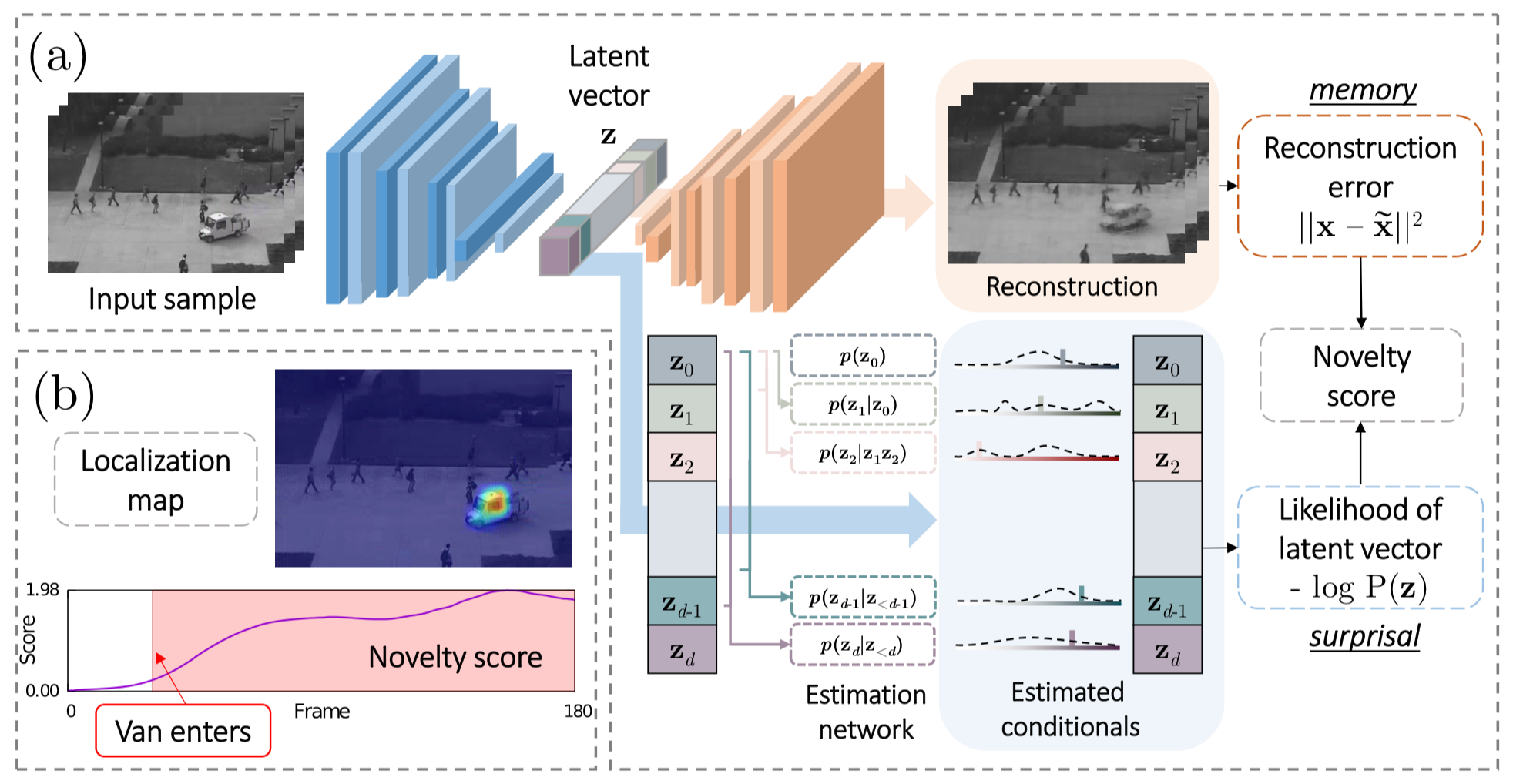}
	\vskip -2.5pt
	\caption{ (a) Network architecture used in latent space auto-regression \cite{AND}. (b) An example of how a novelty score can be used to detect abnormal events.}
	\label{fig:and}
\end{figure}

\noindent \textbf {Latent Space Auto-regression (AND)}\\
The latent space auto-regression framework presented in \cite{AND} learns a representation and models embeddings of the positive class in the latent space simultaneously. \cite{AND} uses an auto-encoder structure to learn the representation. The latent embeddings are modeled using an autoregressive model. Both these objectives are trained end-to-end using the network shown in Figure~\ref{fig:and}(a). 

Let encoder and decoder sub-networks in the auto-encoder be denoted as $\mbox{En}$ and $\mbox{De}$. Given a input ${x}$, latent representation obtained from the encoding process is $\mathbf{z} =\mbox{En}({x})$. The objective of the autoregressive model $h$ is to estimate posterior probability of $\mathbf{z}$ given the positive class $ p(\mathbf{z})$.	

Autoregressive modeling is a technique that predicts sequential outputs based on previous observations. Using this assumption $p(\mathbf{z})$ can be decomposed as:
\begin{equation}
p(\mathbf{z}) = \Pi_{i=1}^m p(\mathbf{z}_i| \mathbf{z}_{<i}),
\end{equation}
where  symbol $<$ denotes  the order of random variables. In \cite{AND}, each conditional probability is modeled using a multinomial with 100 quantization bins. In order to facilitate auto-regression prediction, \cite{AND} proposed a Masked Fully Connected Layer (MFC). It consists of $l$ fully connected layers connected to the latent vector with $j^{th}$ fully connected layer masked, where all rows $<j$ are replaced with zeros. Outputs of each fully connected layer is stacked together to form a $\mathbb{R}^l$ dimensional vector which produces $p(\mathbf{z})$.

The three networks, $\mbox{En}, \mbox{De}$ and $h$ are trained together with the objective of minimizing the reconstruction error $\|\mbox{De}(\mbox{En}({x})- {x}) \|^2$ and the negative log likelihood $- \log(h(\mathbf{z}))$. The full objective function of the network is defined as:
\begin{equation}
\mathcal{L} = \mathbb{E}_{{x}} = \|\mbox{De}(\mbox{En}({x})- {x}) ||^2 -  \lambda  \log(h(\mathbf{z})),
\end{equation}
When this loss is minimized, the network learns to both encode images of the positive class and to model the latent density of positive samples. During inference $\mathcal{L}$ is used to measure the novelty of a query image. As illustrated in Figure~\ref{fig:and}(b), $\mathcal{L}$ can be used to quantify how novel  a given instance is compared to instances in  the positive class. When the value of $\mathcal{L}$ is below a pre-determined threshold it is associated with the positive class. \\

\noindent \textbf {Inter Class Splitting (ICS) }\\
A representation learned on the positive class label will be able to represent majority of data (typical data) well. However, there exists a set of data samples (atypical data) that are not represented well. For an example, if an auto-encoder was chosen to be the representation method, the former set of data points will yield a low reconstruction error since they are represented well. On the other hand, latter data points will yield high reconstruction error. In \cite{schlachter2019}, authors utilize this fact to learn a meaningful representation from one-class data. 

\begin{figure}[!t]
	\centering
	\includegraphics[width=1\linewidth]{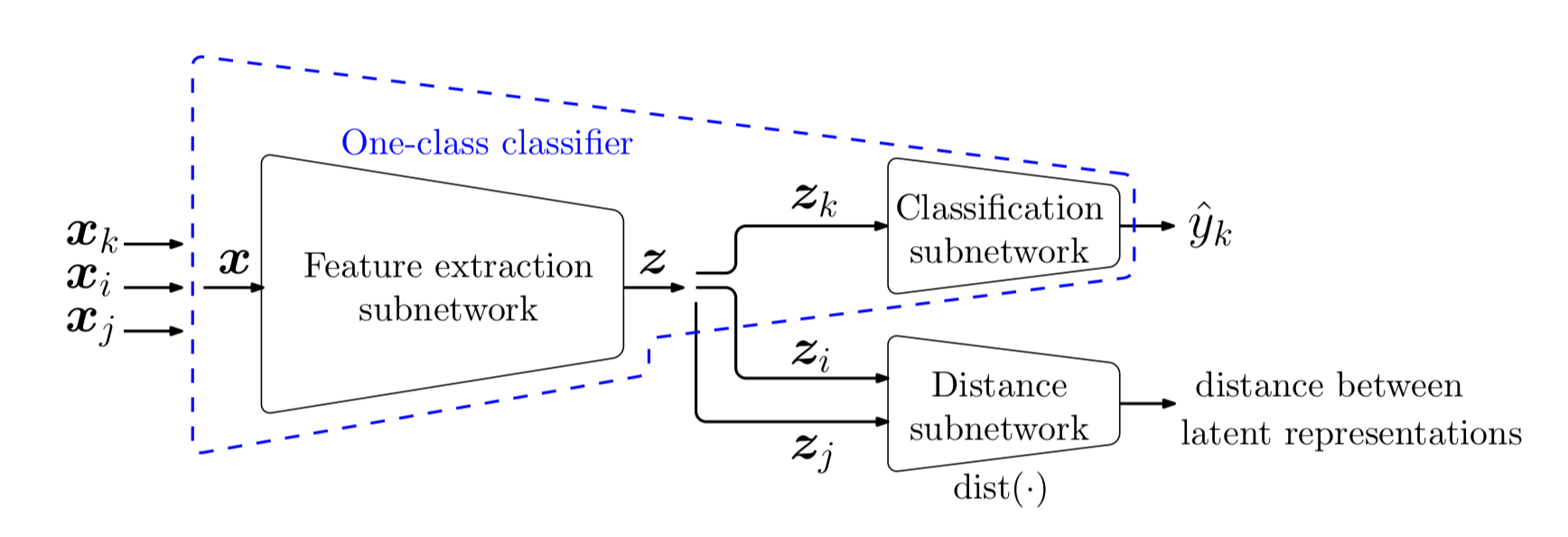}
	\vskip -2.5pt
	\caption{ Network architecture used in to train OCC with inter-class splitting \cite{schlachter2019}.}
	\label{fig:ics}
\end{figure}

%	\begin{figure}[!b]
%		\centering
%		\includegraphics[width=1\linewidth]{hls_oc}
%		\vskip -12.5pt
%		\caption{In hierarchical one-class classifier (HLS-OC), the auto-encoder network is trained with guidance from feature within-class scattered matrix information (WSI-AE), which is then used during inference for one-class classification.}
%		\label{fig:hls_oc}
%	\end{figure}

\begin{figure*}[htp!]
	\centering
	\includegraphics[width=1.00\linewidth]{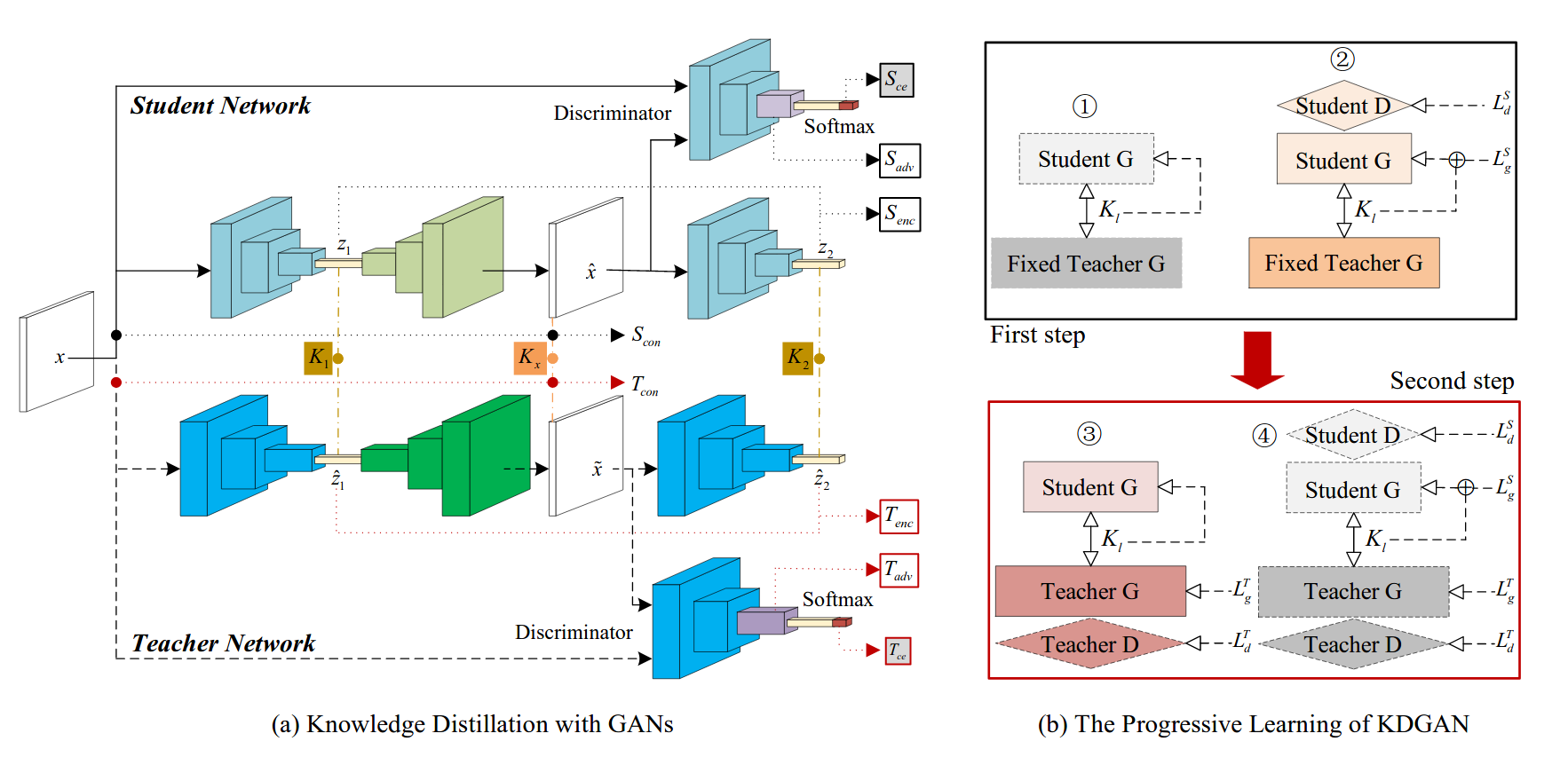}
	\vskip -12.5pt
	\caption{The  flowchart  of  knowledge  distillation  with  GANs  for  one-class  classification \cite{zhangp}. The training involves one teacher and one student network that are trained progressively, where student network is trained with distillation loss to improve its the one-class.}
	\label{fig:p_kdgan}
\end{figure*}

First, an auto-encoder network is trained using all positively labeled data. Then, atypical data samples are identified by considering the reconstruction quality. \cite{schlachter2019} used SSIM as an image quality measure for this purpose.
The core network presented in \cite{schlachter2019} has three components	as shown in Figure~\ref{fig:ics} -- Feature extraction sub-network, Distance sub-network and Classification sub-network. Given an input ${x}$, first a label $y$ is assigned based on whether it is a typical sample or not. For the given input  ${x}$ Feature extraction sub-network produces the corresponding latent representation $\mathbf{z}$. The classification network performs binary classification to classify input into typical or atypical class. The classification network produces binary cross entropy loss $L_{ic}$ defined as: 
\begin{equation}
\mathcal{L}_{ic} = - \frac{1}{N} \sum_{i=1}^{N} [y_i \log \hat{y_i} +  (1-y_i) \log (1-\hat{y_i})],
\end{equation}
where, $\hat{y_i} $ is the prediction produced by the classification network.   Given a pair of data points $({x}_i, {x}_j)$ the distance network generates a score in the range of $[0,1]$ by considering their corresponding latent representations $(\mathbf{z}_i, \mathbf{z}_j)$ . The distance network is trained such that it generates a value closer to $0$ if both images are atypical and to generate a value closer to $1$ when both are typical. The distance network is trained by minimizing the \textit{closeness loss} $L_{cls}$  and dispersion loss $L_{dis}$ defined as:
\begin{equation}
\mathcal{L}_{cls} = - \frac{1}{N} \sum_{i=1}^{N} \log (1- D(\mathbf{z}_{typ, i},    \mathbf{z}_{typ, j \neq i})),
\end{equation}
and,
\begin{equation}
\mathcal{L}_{dis} = - \frac{1}{N} \sum_{i=1}^{N} \log ( D(\mathbf{z}_{atyp, i}, \mathbf{z}_{atyp, j \neq i})),
\end{equation}
where $D(\cdot)$ is a suitable distance measure and $\mathbf{z}_i, \mathbf{z}_i$ are the latent representations of the two input samples considered.  Three sub-networks are trained considering all three losses $ \mathcal{L}_{ic} , \mathcal{L}_{cls} $ and $\mathcal{L}_{dis} $ together. This method encourages to learn an embedding that makes identifying typical samples from atypical samples possible. This is done by minimizing both cross-entropy loss and pair-wise loss in the latent space. Authors show that once the network is trained, the probability corresponding to the typical class can be used to detect positive samples effectively during inference \cite{schlachter2019}.	 \\

\subsubsection{Knowledge Distillation}
A few recent methods have utilized knowledge distillation with student-teacher model training to improve one-class learning.\\
	
\noindent \textbf{Uninformed Students: One-Class Learning (US-OCL)}
The Uninformed Students based One-Class Learning (US-OCL) \cite{bergmann2020uninformed} proposes a student-teacher training strategy for one-class classification. The proposed US-OCL approach, utilizes ensemble of student networks that are trained to regress the output of the teacher network through knowledge distillation. The teacher network is a powerful generative model with an encoder ($\mbox{En}^{teacher}$) and a decoder network ($\mbox{De}^{teacher}$) trained on the one-class training data. The ensemble of student networks then distill the encoder representation using distillation loss expressed as:
\begin{equation}
\begin{array}{c}
\mathcal{L}_{distill}=\|\mbox{En}^{student}({x})-\mbox{En}^{teacher}({x})\|_{2}, \\
\end{array}
\end{equation}
To further improve the quality of the learned representations, US-OCL also introduces a metric learning loss ($\mathcal{L}_{metric}$), i.e. triplets loss \cite{schultz2003learning} computed for a mini-batch, and a compactness loss ($\mathcal{L}_{comp}$), which minimizes the correlations between the samples of a mini-batch. Once the student networks are trained, the errors made by student networks compared to the teacher network prediction is used as scoring function to perform one-class classification.\\

\noindent \textbf{Progressive Knowledge Distillation (P-KDGAN)}\\	
Progressive knowledge distillation (P-KDGAN) \cite{zhangp} proposes one-class method designed to distill knowledge from teacher network and transfer it to a student network through a distillation loss. For this purpose, P-KDGAN trains two GAN networks on given one-class training data, namely student and teacher networks. First, both teacher and student GANs are initialized with the same architecture, with only difference being both architecture have different number of channels. Both the networks are trained with the reconstruction loss $S_{con}$, latent space loss $S_{enc}$ and adversarial loss $S_{adv}$ defined as follows:
\begin{equation}
\begin{array}{c}
S_{con}=\|{x}-\hat{{x}}\|_{1} \\
S_{enc}=\left\|\mathbf{z}_{1}-\mathbf{z}_{2}\right\|_{2} \\
S_{adv}=\|\mbox{En}({x})-\mbox{En}(\hat{{x}})\|_{2} \\
L^S_{g}=w_{con} S_{con}+w_{enc} S_{enc}+w_{adv} S_{adv},
\end{array}
\end{equation}
where, $w_{con}$, $w_{enc}$ and $w_{adv}$ are weights for the corresponding loss functions. The same loss is used for training the teacher network denoted as $L^T_{g}$. To distill the knowledge from the student network, distillation losses $K_1$, $K_x$ and $K_2$ as shown in the Fig.~\ref{fig:p_kdgan}. The losses $K_1$ and $K_2$ minimize the distance between student and teacher representations learned by an encoder and discriminator network through L2 error and $K_x$ minimizes the L1 error in the reconstructed input images by student and teacher network. Both student and teacher networks are trained using a progressive training strategy with two steps. In the first step, the student network is trained with distillation losses and a generator loss with fixed teacher parameters. In second step, both student and teacher networks are trained with generator loss and student network is fine-tuned with a distillation loss as well.

\section{Datasets and Evaluation Protocols}\label{sec:data_and_metrics}
In this section we present a discussion of commonly used datasets and evaluation metrics for OCC.

\subsection{Multi-class Benchmark Datasets}
Image based OCC performance is commonly benchmarked using standard multi-class datasets. During evaluation, each class present in the dataset is considered one at a time as the positive class. The model is trained using the training split of the positive class. The trained model is tested against the test split of all classes. Here, all other classes are considered to be negative during testing. We list datasets used commonly in recent literature below.\\ 

\noindent  \textbf{COIL100 :} COIL100 \cite{nene1996columbia} is a multi-class dataset where each object class data is captured in  different poses. There are 100 image classes in the dataset with a few images per class (typically less than hundred).\\

\noindent  \textbf{MNIST :} The MNIST  dataset \cite{nene1996columbia} contains $28\times28$ hand-written digits from 0-9.  \\

\noindent  \textbf{fMNIST :} fMNIST \cite{xiao2017fmnist} is intended to be a replacement for MNIST, where the dataset comprises of 28$\times$28 images of fashion apparels/accessories. \\

\noindent  \textbf{CIFAR10 :} CIFAR10 \cite{krizhevsky2010cifar} is an object recognition dataset that consists of images from 10 classes. It contains images from four vehicle classes and six animal classes. All images are centered and cropped.\\

\subsection{Abnormal 1001 Dataset}
\begin{figure}[htp!]
	\centering
	\includegraphics[width=.8\linewidth]{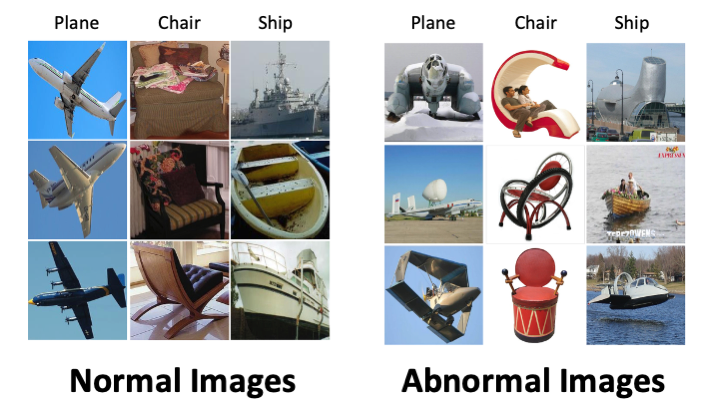}
	\caption{ Sample images from the Abnormal 1001 (abnormal images) \cite{Saleh:2013:OAD:2514950.2516141} and PASCAL (normal images) datasets \cite{Everingham10}.}
	\label{fig:samples}
\end{figure}

The 1001 Abnormal Objects Dataset \cite{Saleh:2013:OAD:2514950.2516141} contains 1001 abnormal images belonging to six classes which are originally found in the PASCAL dataset \cite{Everingham10}.  Six classes considered in the dataset are Airplane, Boat, Car, Chair, Motorbike and Sofa. Each class has at least one hundred abnormal images in the dataset. A sample of abnormal images and the corresponding normal images from the PASCAL dataset are show in Figure~\ref{fig:samples}. Abnormality of images has been judged based on human responses received on the Amazon Mechanical Turk. For each of the considered object classes,  a model is trained using training images from the PASCAL dataset. During testing, test images of same class from the PASCAL dataset is considered to be positive data. Negative class data are sampled from the corresponding class of the Abnormal1001 dataset.

\subsection{MVTec-AD Dataset}
\begin{figure}[htp!]
	\centering
	\includegraphics[width=.8\linewidth]{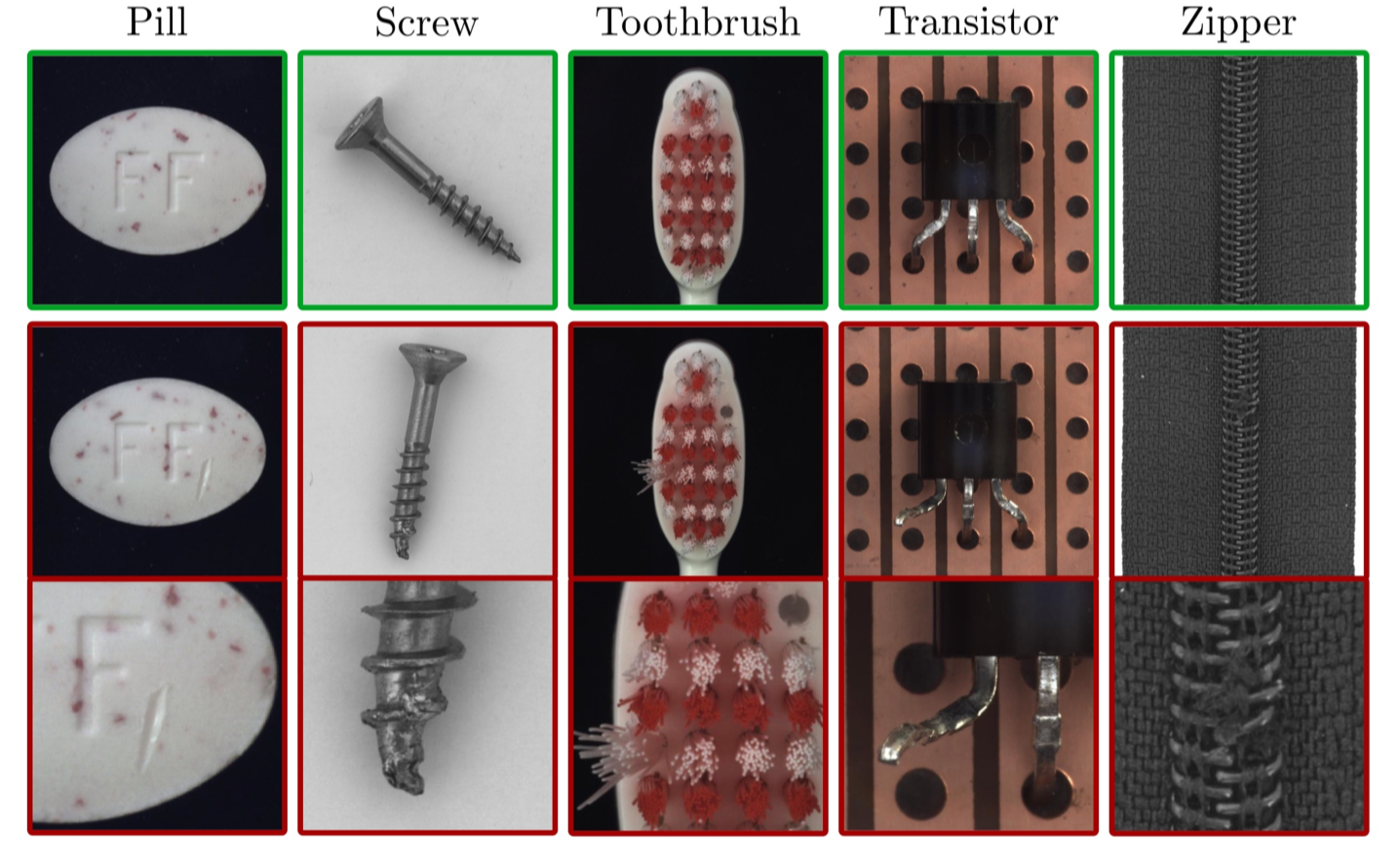}
	\caption{ Sample images from MVTec--AD dataset \cite{Bergmann_2019_CVPR}. Top: normal images from different categories. Middle: abnormal images from corresponding categories. Bottom: magnified images showing defects in abnormal images.}
	\label{fig:anomaldata}
\end{figure}

The MVTec-AD dataset \cite{Bergmann_2019_CVPR} contains image classes that are typically found in an industrial setting. It contains 5 classes belonging to textures (Carpet, Grid, Leather, Tile and Wood) and 10 classes belonging to objects (Bottle, Cable, Capsule, Hazelnut, Metal, Nut, Pill, Screw, Toothbrush, Transistor and Zipper). Authors have explicitly identified \textit{normal} and \textit{abnormal} images for each class as shown in Figure~\ref{fig:anomaldata}. Identified \textit{abnormal} images  contain product defects or natural wear and tear. The dataset contains 3629 training images. It contains 467 \textit{normal} images and 1258 \textit{abnormal} images  in the test set. Performance of OCC can be evaluated by first training a one class model for each class, where  \textit{normal} images are treated as positive data. The trained model is tested on both \textit{normal} and \textit{abnormal} images of the same class, where \textit{abnormal} images are treated as negative data. 

\subsection{Evaluation Metrics}
One class classifiers are evaluated using a test dataset that comprises of positive data and negative data. Therefore, testing procedure of one class classifiers are analogous to binary classifiers / detectors. Majority of previous works in the literature \cite{HOFFMANN2007863}, \cite{ocfeatures}, \cite{rotations}, \cite{AND}, \cite{ocgan}  have used Receiver Operating Characteristics (ROC) curve to report performance of one class classification. The ROC curve represents the relationship between the false positive rate (FPR) and the true positive rate (TPR) obtained by a classifier. They are defined as:
\begin{equation}
TPR = \frac{\text{Number of correctly classified positive samples}}{\text{Number of positive samples}},
\end{equation}
and
\begin{equation}
FPR = \frac{\text{Number of misclassified negative samples}}{\text{Number of negative samples}},
\end{equation}

The following metrics can be derived based on the ROC curve:\\

\noindent \textbf{AUC-ROC curve}. Area under the ROC curve is a commonly used metric to evaluate the effectiveness of OCC methods. Ideally, the area should account for 1.0. In the worst case, prediction will be equal to random guessing when the AUC-ROC is 0.5. AUC-ROC does not demand of using any specific operating points when the metric is evaluated. Therefore, it captures the effectiveness of the method independent of an operating point. \\

\noindent \textbf{FPR @ TPR}. This metric reports FPR at a selected TPR value from the ROC curve. Equal error rate (EER) is a special case of this metric. EER is the FPR value reported at the point in the ROC curve when both FPR and TPR are equal.\\

\noindent \textbf{Half Total Error Rate (HTER) @ TPR}. Half total error rate quantifies the probability of obtaining a wrong prediction given probability of observing a positive and a negative query is equal. This is quantified as:
\begin{equation}
HTER = \frac{FPR+(1-TPR)}{2},
\end{equation}
This metric is commonly used in biometrics applications. It is the usual practice to report HTER when FPR and TPR are equal. 

\section{Discussion and future research directions}\label{sec:openproblems}
This article attempted to provide an overview of recent developments in  OCC  for  visual recognition. One class classification promises  to  be  an  active  area  of  research.  Based on the analysis and results of various methods and the trend of other developments in computer vision, we believe that deep networks will dominate further research in the field of OCC.  We make the following observations regarding future trends in research in OCC.\\

\noindent \textbf{OCC in complex visual object datasets.} The majority of previous works in the literature (in exception of \cite{rotations})  have carried out evaluations on relatively simpler datasets that contain low amount of details and variations. Based on the reported performance it can be seen that although the existing methods perform very well in simpler datasets (such as MNIST, COIL100), they fail to generalize well to more complex datasets (such as CIFAR10). In multiple-class classification CIFAR10 is no longer considered as a challenging dataset. %It has moved to more challenging datasets such as ImageNet and Places365. 
One of the open problems in OCC is to create sophisticated solutions that generalize well to more complex datasets.\\

\noindent \textbf{Semi-supervised OCC.} Conventionally OCC methods assume the availability of only positively labeled data during training. However, in practical applications, this limitation does not always hold. Semi-supervised OCC studies how to find solutions in the presence of additional annotated data. Semi-supervised OCC has been formulated in a couple of ways.  In the first case, the availability of labeled data from an OOD dataset is assumed \cite{ocfeatures}, \cite{oza2019one}, \cite{masana2018metric}. In the other case, the availability of a few annotated data from positive and negative classes are assumed   \cite{Ruff2020Deep}, \cite{akcay2018ganomaly}.  More semi-supervised OCC methods are needed that can leverage additional datasets to improve the performance.  \\

\noindent \textbf{One-Class Adversarial Attacks.} Another limitation of current OCC methods is that they can not deal with adversarial data.  It is a well-known fact that carefully designed imperceptible perturbations can be used to fool a deep learning-based model to make incorrect predictions \cite{szegedy2013intriguing}, \cite{goodfellow2014explaining}, \cite{carlini2017towards}.   These types of adversarial attacks are easy to deploy and may be encountered in real-world applications \cite{amodei2016concrete}.  However, most of the work studying adversarial attacks and defense strategies focus on multi-class classification models \cite{goodfellow2014explaining}, \cite{carlini2017towards}, \cite{madry2018towards}, \cite{xie2019feature}.  Recent work \cite{salehi2020arae} showed that a deep network-based OCC approach is also vulnerable to adversarial attacks.  However more is needed.   Extensive study in understanding how adversarial attacks affect one-class models will help in creating more secure models for real-world applications. \\
	
\noindent \textbf{Domain generalizable OCC.} When a one class classifier is trained, it is done with the assumption that the query samples observed by the classifier during inference would follow the same distribution as the training images. However, this assumption may not hold in practice. A considerable amount of interest has emerged in recent multi-class classification literature in producing classifiers that generalize well across different image distributions (domains) \cite{ganin15}, \cite{coral}, \cite{dcoral}, \cite{Patel_DA_SPM}. However, this area of research is largely unexplored for OCC \cite{sindagi}, \cite{chen2014transfer}. Therefore, developing techniques that produce  generalizable OCC  across different image distributions remains an open problem.\\

\noindent \textbf{One-Class Neural Architecture Search.} In recent years, there has been a growing interest in automating architecture design/engineering using Neural architecture search (NAS) methods \cite{elsken2019neural}.  Given different basic convolution operations, the idea is to learn the connections in feature search cell using NAS methods to obtain an optimal network for one-class classification. More is needed to explore the possibility of using NAS for one-class classification.\\

%Neural architecture search (NAS) has shown to be really helpful in finding optimal architecture design for any given task \cite{elsken2019neural}. Many have utilized it to develop the most effective network architecture for tasks such as classification \cite{pham2018efficient}, \cite{liu2018progressive}, segmentation \cite{liu2019auto}, \cite{weng2019unet}, object detection \cite{chen2019detnas}, \cite{wang2020fcos} etc. These architecture search techniques are yet to be studied for the one-class classification networks. Learning optimal architecture for one-class models have a good potential to improve the performance. Hence, development of NAS techniques specifically for the task of one-class classification is an interesting direction for the future research.  \\

\noindent \textbf{Explainable OCC.} In past decade, the majority of computer vision algorithms have transitioned to deep convolutional neural network (DCNN) based models. However, due to the complexity of DCNN models, the decisions they make are hard to interpret \cite{lipton2018mythos}. Hence, explaining a DCNN is essential towards developing transparent models that not only make decisions, but can also explain why the decision was made. Many works have contributed in developing more transparent models, but they all focus on recognition tasks \cite{selvaraju2017grad}, \cite{chattopadhay2018grad}, \cite{fukui2019attention}, \cite{mahendran2015understanding}, \cite{zeiler2014visualizing}. Since all recently proposed OCC models also work on deep convolutional neural networks, a special focus on explaining OCC models is critical. There are few papers addressing this issue for OCC models \cite{liu2020towards}, \cite{liznerski2020explainable}, \cite{nguyen2019gee}, providing first steps towards developing more transparent OCC models. However, explainable OCC models are still largely unexplored and more work in this direction is needed.\\

\noindent \textbf{Federated learning for OCC.} Training computer vision models with the help of decentralized data has been an interesting problem that is gaining attention in recent years. Federated learning \cite{mcmahan2017communication} was proposed to specifically address this issue to train object recognition models with the help of decentralized data. Most of the federated learning methods are proposed only for multi-class classification models \cite{konevcny2016federated}, \cite{smith2017federated}, \cite{li2020federated}. However, these multi-class federated learning approaches are not directly applicable to train one-class classification methods with decentralized data. This is due to the additional challenges posed by the nature of one-class classification problem. For example, in federated multi-class classification, homogeneous distribution of multiple-category across individual data centers is assumed, which is not possible for the federated one-class setting. Currently there are no methods proposed to train one-class classifiers with decentralized data, making it an interesting direction for future work.

{\small
	\bibliographystyle{ieee}
	\bibliography{egbib}
}

\end{document}